%% file: ijcai25.tex
\documentclass{article}
\usepackage{ijcai25}

\usepackage{times}
\usepackage{soul}
\usepackage{url}
\usepackage[hidelinks]{hyperref}
\usepackage[utf8]{inputenc}
\usepackage[small]{caption}
\usepackage{graphicx}
\usepackage{amsmath}
\usepackage{afterpage}
\usepackage{dblfloatfix}
\usepackage{amsthm}
\usepackage{booktabs}
\usepackage{algorithm}
\usepackage{algorithmic}
\usepackage{pifont}
\usepackage{bbm}
\usepackage{amsmath,amsfonts}
\usepackage{algorithmic}
\usepackage{color}
\usepackage{multirow}
\usepackage{multicol}
\usepackage{colortbl}
\usepackage{subfigure}
\usepackage[table]{xcolor}
\usepackage{float}
\definecolor{grey}{HTML}{DCDCDC}
\usepackage[switch]{lineno}

\usepackage{xspace}
\newcommand{\ourmethod}{{MbaGCN}\xspace}

\title{
\begin{tabular}{@{}c@{}}
\multirow{1}{*}{
\vspace{10cm}\includegraphics[width=0.5cm, height=0.5cm, clip=False]{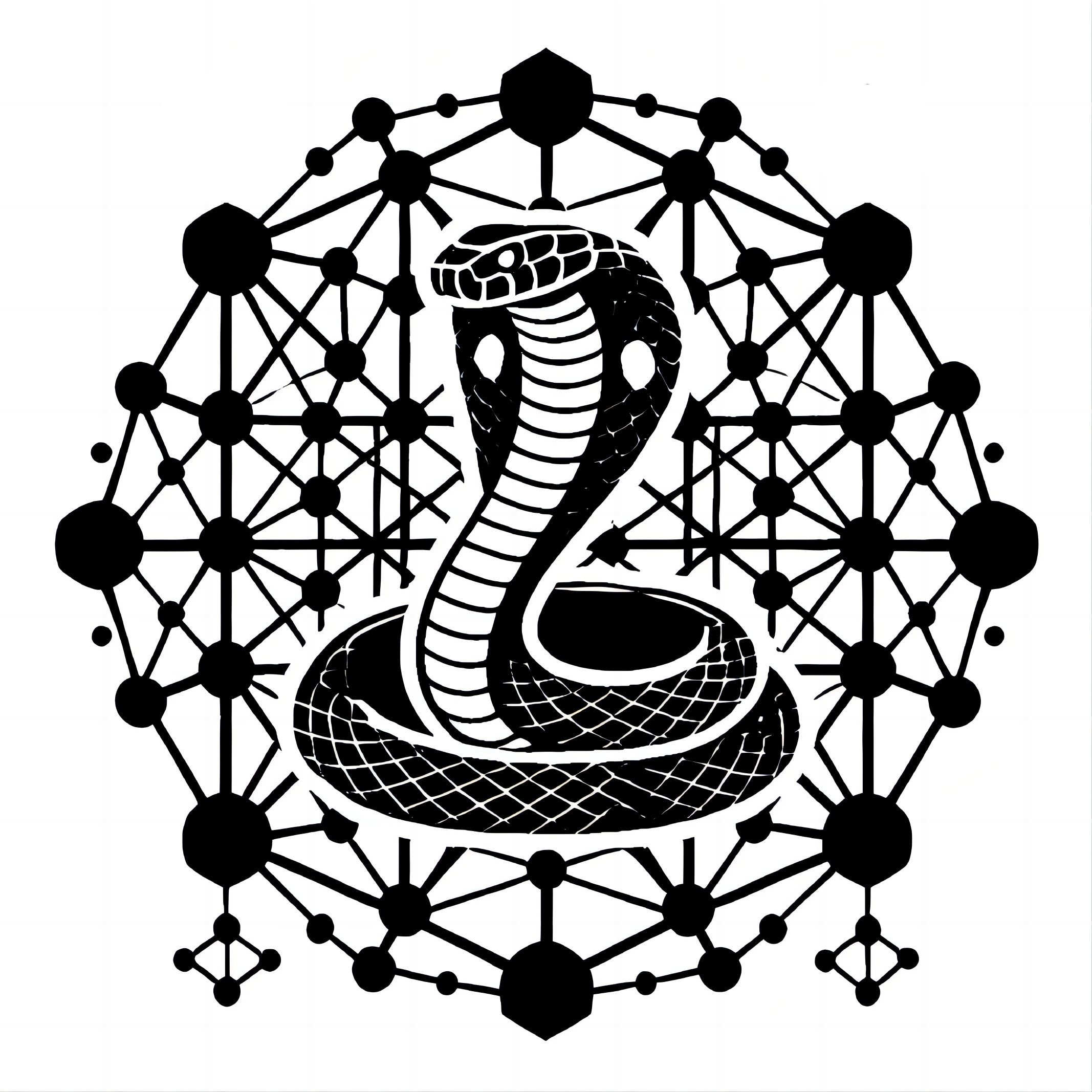} 
}
\end{tabular}Mamba-Based Graph Convolutional Networks: \\Tackling Over-smoothing with Selective State Space}

\author{
Xin He$^1$
\and
Yili Wang$^1$\and
Wenqi Fan$^{2}$\and
Xu Shen$^1$\and
Xin Juan$^1$\and
Rui Miao$^1$\And
Xin Wang$^1$\\
\affiliations
$^1$Jilin University\\
$^2$The Hong Kong Polytechnic University\\
\emails
\{hexin20, wangyl21\}@mails.jlu.edu.cn,
wenqifan03@gmail.com,
\{shenxu23, junxin22, ruimiao20\}@mails.jlu.edu.cn,
xinwang@jlu.edu.cn,
}

\begin{document}

\maketitle
\begin{abstract}
Graph Neural Networks (GNNs) have shown great success in various graph-based learning tasks. 
However, it often faces the issue of over-smoothing as the model depth increases, which causes all node representations to converge to a single value and become indistinguishable.
This issue stems from the inherent limitations of GNNs, which struggle to distinguish the importance of information from different neighborhoods.
In this paper, we introduce \ourmethod, a novel graph convolutional architecture that draws inspiration from the Mamba paradigm—originally designed for sequence modeling. 
\ourmethod presents a new backbone for GNNs, consisting of three key components: the \textbf{Message Aggregation Layer}, the \textbf{Selective State Space Transition Layer}, and the \textbf{Node State Prediction Layer}. 
These components work in tandem to adaptively aggregate neighborhood information, providing greater flexibility and scalability for deep GNN models. 
While \ourmethod may not consistently outperform all existing methods on each dataset, it provides a foundational framework that demonstrates the effective integration of the Mamba paradigm into graph representation learning. Through extensive experiments on benchmark datasets, we demonstrate that \ourmethod paves the way for future advancements in graph neural network research. 
Our code is in \href{https://github.com/hexin5515/MbaGCN}{https://github.com/hexin5515/MbaGCN}.
\end{abstract}
\input{sections/introduction}
\input{sections/relatedwork}
\input{sections/preliminary}

\input{sections/method}
\input{sections/experiment}
\input{sections/conclution}
\input{sections/acknowledgment}
\input{sections/contributionstatement}

\bibliographystyle{named}
\bibliography{ijcai25}

\clearpage
\input{sections/appendix}

\end{document}

%% file: sections/introduction.tex
\section{Introduction}

\begin{figure}[t!]
  \includegraphics[width=1\linewidth]{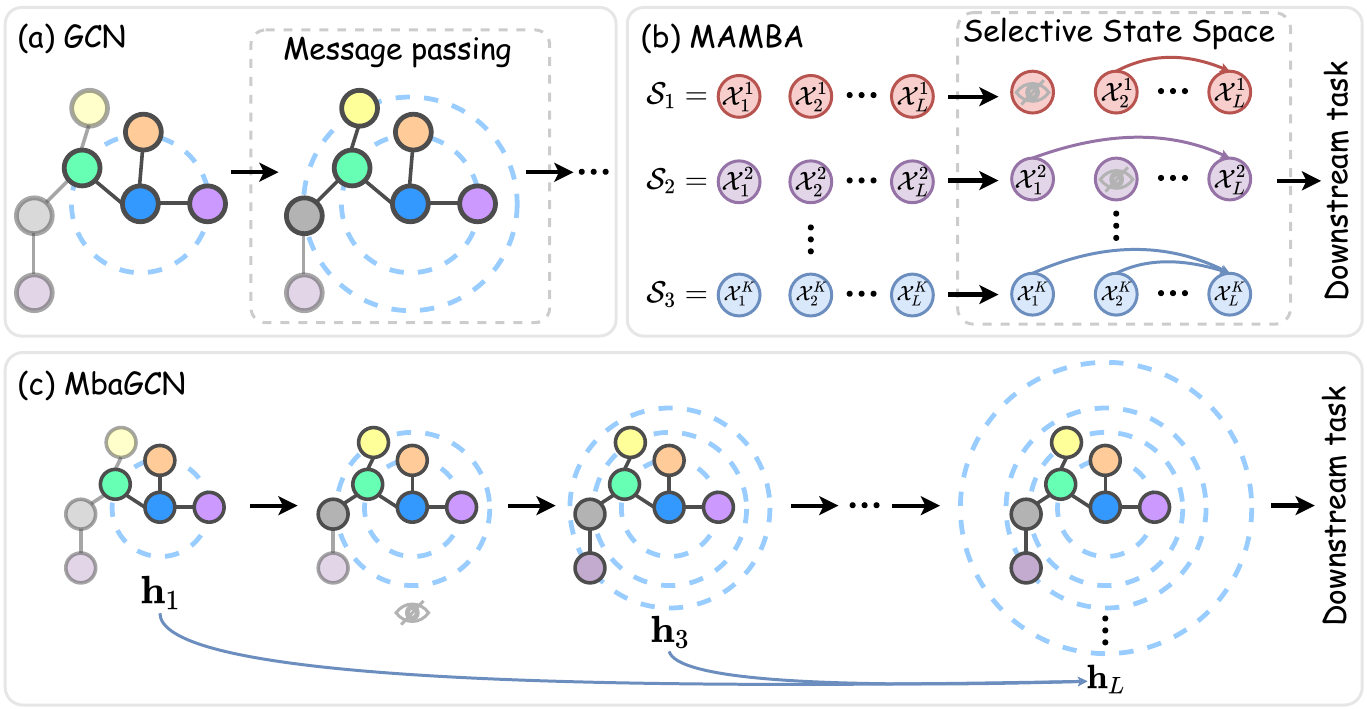}
  \caption{Comparison of GCN, MAMBA, and MbaGCN.}
  \label{fig:motivation}
\end{figure}

In recent years, Graph Neural Networks (GNNs)~\cite{zhang2021nested} have gained significant attention for their ability to process graph data, achieving success in node classification~\cite{shen2024optimizing,shen2024graph}, recommendation systems~\cite{qin2024learning}, and biology~\cite{shen2025raising,GraphSAM}. Among these, Graph Convolutional Networks (GCNs)~\cite{wang2024unifying,wang2022adagcl} stand out due to their capability to propagate node features through a graph's topology and extract knowledge from non-Euclidean spaces (see Fig.\ref{fig:motivation}(a)). 
By using convolution operators, GCNs aggregate information from neighboring nodes, enabling effective learning for tasks such as link prediction and protein interaction analysis.
However, GNNs face a key limitation: 
their struggle to effectively differentiate the significance of information coming from nodes located at different distances within the graph. This limitation directly leads to the issue of over-smoothing in GNNs. In other words, node representations gradually become more similar to each other as the depth of the network increases, limiting the scalability and performance of the GNNs.

Mamba~\cite{gu2023mamba,dao2024transformers}, originally designed for sequence modeling, addresses a fundamental challenge in sequence-based tasks — the difficulty of capturing long-range dependencies. It incorporates a \textbf{selective state space mechanism} (see Fig.\ref{fig:motivation}(b)) that dynamically and adaptively compresses information from nodes at different distances, retaining only the most relevant data for the downstream tasks. This is crucial for tasks such as language modeling~\cite{wang2021structure,yuan2024instance} or time-series forecasting~\cite{grazzi2024mamba,wang2025mamba}, where the importance of information diminishes with distance. 
The key insight of Mamba’s selective state space mechanism lies in its ability to differentiate the relevance of information at various distances~\cite{zhu2024vision}. 
Therefore, Mamba is inherently suited to address the over-smoothing problem in graph data, where nodes at different hops should contribute varying levels of importance during aggregation. 
Instead of uniformly aggregating information from all neighboring nodes, the Mamba-based approach adaptively aggregates based on each neighborhood's relevance. This helps retain key features from different-order neighborhoods, mitigating over-smoothing and improving the GCNs' ability to capture multi-hop relationships (see Fig.\ref{fig:motivation}(c)). 

In this work, we propose a novel architecture that integrates the Mamba paradigm into GNNs, named \textbf{M}am\textbf{ba}-based \textbf{G}raph \textbf{C}onvolutional \textbf{N}etwork (\textbf{\ourmethod}). Inspired by Mamba’s selective state space model, \ourmethod aims to address the limitations of traditional GCNs by adaptively compressing and propagating node features. Specifically, it preserves only the most relevant information for downstream tasks, enabling more effective aggregation and propagation. 

\ourmethod consists of three key components: the \textbf{Message Aggregation Layer (MAL)}, the \textbf{Selective State Space Transition Layer (S3TL)}, and the \textbf{Node State Prediction Layer (NSPL)}. The MAL performs a simple message-passing operation that aggregates neighborhood information, helping nodes incorporate information from their neighbors. The S3TL introduces a selective state space mechanism that identifies and retains the most important neighborhood features, condensing the graph’s information into state vectors. This ensures that relevant node features are preserved, while redundant or less useful data is discarded. NSPL regulates the information flow within the same-order neighborhood, ensuring that essential local features are maintained while allowing the model to consider the global context. By alternating between these layers, \ourmethod balances local and global information propagation, adapting the information flow to the graph structure. The combination of MAL, S3TL and NSPL allows \ourmethod to scale effectively with deeper architectures and complex graph data, offering a promising solution to the challenges faced by traditional GCNs.


The contributions of this work are summarized as follows:

\begin{itemize}
    \item We introduce a new approach to integrate the Mamba paradigm into GNNs, using its selective state space mechanism to address over-smoothing in graph representation learning.
    \item We propose \ourmethod, a new graph convolutional architecture that alternates between the MAL and the S3TL to adaptively retain important information from neighborhoods of different orders, while the NSPL refines the learned node representations.
    \item \ourmethod improves information flow through deeper GNN architectures by selectively retaining important features from neighborhoods, enabling the model to capture both local and global graph structures better.
    \item Our experiments on benchmark datasets demonstrate the potential of \ourmethod and provide valuable insights for future research directions in GNN development.
\end{itemize}

%% file: sections/relatedwork.tex
\section{Related Wrok}

\subsection{Over-smoothing on Graph Data}
Over-smoothing in graph representation learning arises as the number of GCN layers increases (repeated application of Laplacian smoothing)~\cite{li2018deeper,zhang2021node,rusch2023survey}.
This leads to the convergence of all node representations within the same connected component of the graph to a single value, severely impacting the model's performance~\cite{wu2024demystifying,zhai2024bregman}.
In recent years, researchers have proposed various solutions to address this issue from different perspectives.
For instance, introducing residual connections between layers preserves the integrity of node representations~\cite{chen2020simple,zhu2021simple}, employing personalized neighborhood aggregation based on PageRank better captures important node features~\cite{chien2020adaptive}, and applying regularization techniques that incorporate both graph structure and node features mitigates over-smoothing~\cite{yan2022two,miao2024rethinking}.
These methods collectively enhance the model’s ability to retain meaningful node information across deeper layers and effectively mitigate the over-smoothing issue.
Recent studies~\cite{lieber2024jamba,yang2024plainmamba,hu2025zigma} show that Mamba can filter out irrelevant information from long sequential data, which inspires us to propose a Mamba-based GCN. This approach adaptively aggregates neighborhood information, mitigating over-smoothing and improving scalability in deeper models.

\subsection{Mamba \& Mamba with Graph}
Mamba~\cite{gu2023mamba,dao2024transformers} was originally designed for sequence modeling that selectively filters and compresses information to retain only the most relevant data.
Research on Mamba is still in its early stages, but some works have already focused on using Mamba to process graph data.
Such as using Mamba in series with GCN improves the prediction of patients' health status~\cite{tang2023modeling}, combining Mamba and GCN in parallel overcomes GCN’s limitation in capturing long-range dependencies between distant nodes~\cite{wang2024graph,behrouz2024graph}, and applying Mamba directly captures long-distance dependencies between nodes in the graph~\cite{ding2024recurrent}.
\textbf{These methods primarily combine the independent Mamba and GCN modules in various ways, yet they fail to fully harness Mamba's capabilities in graph-structured data processing.}
Therefore, in this paper, we propose a Mamba-based Graph Convolutional Network (\ourmethod) that integrates Mamba into the graph convolution process in a more cohesive manner, effectively addressing the over-smoothing problem in graph representation learning.

%% file: sections/preliminary.tex
\section{Preliminary and Background}

\begin{figure*}[ht!]
  \includegraphics[width=\textwidth]{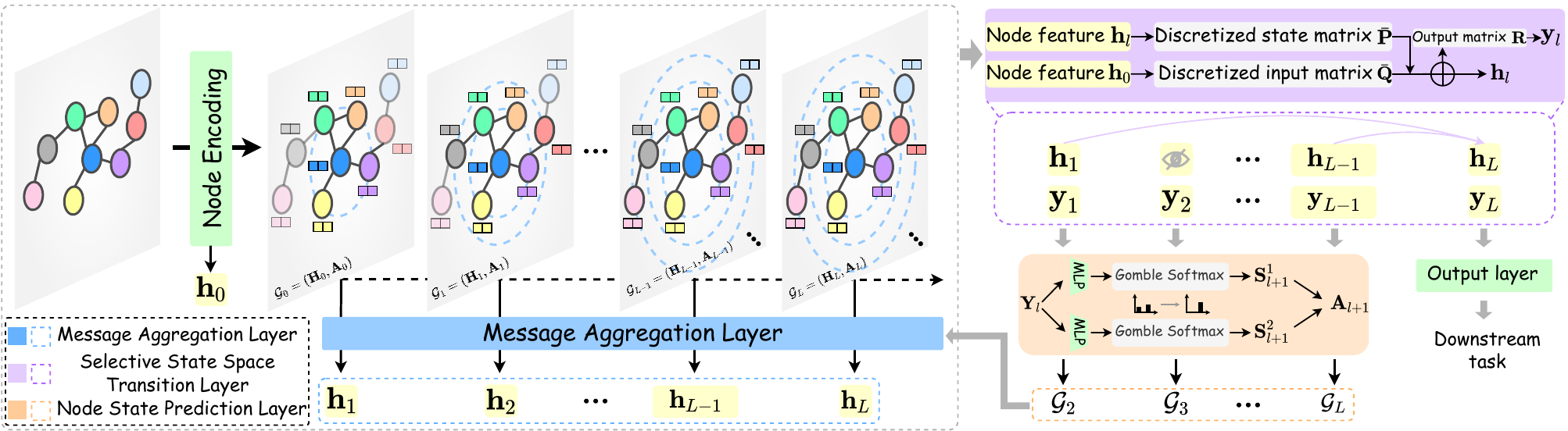}
  \caption{The Framework of \ourmethod. }
  \label{fig:framework}
\end{figure*}

\subsection{Mamba}
Mamba~\cite{gu2023mamba} is a class of linear time-varying systems that map an input sequence $x(t) \in \mathbb{R}^{L}$ to an output sequence 
$y(t) \in \mathbb{R}^{L}$, utilizing a latent state vector $h(t) \in \mathbb{R}^{N \times L}$, a state matrix $\mathbf{P}\in \mathbb{R}^{N\times N}$, an input matrix $\mathbf{Q}\in \mathbb{R}^{N\times 1}$, and an output matrix $\mathbf{R}\in \mathbb{R}^{1\times N}$. The relationship between these components is given by the following equations\footnote{In the graph domain, the matrix \( \mathbf{A} \) has a special meaning, so we modify the notation from Mamba: \( \mathbf{A} \rightarrow \mathbf{P} \), \( \mathbf{B} \rightarrow \mathbf{Q} \), and \( \mathbf{C} \rightarrow \mathbf{R} \).}:


\begin{equation}
	\begin{aligned}
	  h'(t)&=\mathbf{P}h(t)+\mathbf{Q}x(t), \\
    y(t)&=\mathbf{R}h(t).
	\end{aligned}
\end{equation}

Due to the challenges in solving the above equation within the deep learning paradigm, the discrete space state model~\cite{gu2021efficiently} introduces additional parameter $\mathbf{\Delta}$ to discretize the aforementioned system, which can be formulated as follows:
\begin{equation}
	\begin{aligned}
    h(t)&=\bar{\mathbf{P}}h(t)+\bar{\mathbf{Q}}x(t)\\
    y(t)&=\mathbf{R}h(t)
\end{aligned}
\end{equation}
where
\begin{equation}
	\begin{aligned}
    \bar{\mathbf{P}}&={\rm exp}(\mathbf{\Delta} \mathbf{P})\\
    \bar{\mathbf{Q}}&=(\mathbf{\Delta} \mathbf{P})^{-1}({\rm exp}(\mathbf{\Delta} \mathbf{P})-\mathbf{I})\cdot \mathbf{\Delta} \mathbf{Q}
\end{aligned}
\end{equation}
where $\bar{\mathbf{P}}$ and $\bar{\mathbf{Q}}$ are the discrete state matrix and discrete input matrix, ${\rm exp}(\cdot)$ refers to the exponential function with base $e$. 
On this foundation, Mamba further introduces a data-dependent state transition mechanism, which generates unique  $\mathbf{P}$, $\mathbf{Q}$, $\mathbf{R}$, and  $\mathbf{\Delta}$ based on the input data, thereby achieving outstanding performance in language modeling. In this work, we adapt Mamba's core ideas to construct a novel graph convolution paradigm that adaptively aggregates information from nodes across varying neighborhood orders. This adaptive aggregation helps address the issues of over-smoothing 
in deeper graph networks, which is a common challenge in traditional graph representation learning.

\subsection{Problem Statement}
Given an undirected graph $\mathcal{G} = (\mathcal{V}, \mathcal{E})$ with $N$ nodes and $M$ edges, where $\mathcal{V}$ is the set of $N$ nodes and $\mathcal{E}$ is the set of $M$ edges. We define the adjacency matrix as $\mathbf{A} \in \mathbb{R}^{N \times N}$, where $\mathbf{A}_{ij} = 1$ if there is an edge between node $v_i$ and node $v_j$, and $\mathbf{A}_{ij} = 0$ otherwise.
We also define the node feature matrix as $\mathbf{X} \in \mathbb{R}^{N \times d}$, which contains a $d$-dimensional feature vector for each node. In fully supervised node classification tasks, \ourmethod aims to optimize its parameters using the given training set of labeled samples and their corresponding ground truth $\mathbf{Y} \in \mathbb{R}^{N \times c}$ with the number of classes $c$, enabling the model to achieve better performance than shallow models even when the depth is increased to aggregate higher-order neighbors.

%% file: sections/method.tex
\section{MbaGCN}
The structure of \ourmethod, shown in Fig.\ref{fig:framework}, includes the Message Aggregation Layer (MAL), Selective State Space Transition Layers (S3TL), and Node State Prediction Layer (NSPL). The MAL aggregates neighborhood information using a basic graph aggregation operation. The S3TL fuses neighborhood features with node intrinsic features via a spatial state transition equation, condensing them into a state vector for the next iteration.
\ourmethod alternates between MAL and S3TL, strategically modulating the influence of information from neighborhoods of varying orders within the node features. This approach effectively aligns the iterative Mamba paradigm with unordered graph structures, addressing the over-smoothing issue. Additionally, the NSPL is positioned between the MAL and S3TL, which regulates the information flow within the same-order neighborhood. This helps further refine the node feature representation.


\subsection{Alternating MAL and S3TL}

Graph representation learning~\cite{xu2021self} relies on effectively aggregating neighborhood information to enhance node representations. However, traditional methods~\cite{wan2021contrastive,li2021dual} often face challenges such as over-smoothing, where node-specific features become indistinguishable as the network deepens. To address this, we introduce an alternating mechanism between the MAL and the S3TL. This alternating process refines the aggregation of neighborhood information, with each layer alternating between capturing local feature details and adaptively compressing relevant neighborhood information. 

\subsubsection{Message Aggregation Layers (MAL)}

The MAL is the initial stage of \ourmethod, responsible for capturing local feature information by aggregating neighboring node features. It serves as the foundation for the feature refinement process, enabling effective aggregation of node information from the graph’s local structure. The MAL is computed as follows:
\begin{equation}
	\begin{aligned}
 \label{eq:mal}
    \textbf{H}_{l} = \Tilde{\mathbf{D}}^{-\frac{1}{2} } \mathbf{A} \Tilde{\mathbf{D}}^{-\frac{1}{2}} \mathbf{H}_{l-1},
\end{aligned}
\end{equation}
where $\Tilde{\mathbf{D}}$ is the diagonal degree matrix, and $\mathbf{A}$ is the adjacency matrix. 
$\mathbf{H}_{l-1}$ and $\mathbf{H}_{l}$ represent the feature matrices of node after aggregating $(l-1)$-hop and $l$-hop neighborhoods, respectively.
The MAL efficiently aggregates the features of a node’s neighbors, yet it struggles to scale with deeper networks or more complex graph structures. This necessitates the introduction of the S3TL, which utilizes the selective state space model to address this issue.

\subsubsection{Selective State Space Transition Layers (S3TL)}
Traditional neighborhood aggregation methods~\cite{jin2021node,yu2022multiplex} often struggle to capture the full range of important information, especially when processing nodes at varying distances. To effectively aggregate information from diverse neighborhood orders without losing relevant features, S3TL adaptively combines neighborhood information with node-specific features. This process leverages a selective state space transition mechanism that compresses the aggregated data, retaining only the most pertinent information. 
At the same time, redundant information is discarded, improving the model's adaptability to deeper layers and more complex graph structures.

To effectively handle the diverse structures and features of nodes, it is crucial that the fusion process in S3TL can adapt accordingly. To achieve this, we introduce the input-related approach~\cite{gu2023mamba} for generating the input matrix $\mathbf{Q}$, output matrix $\mathbf{R}$, and additional parameter matrix $\mathbf{\Delta}$, which are essential for the selective state space transition. These matrices are computed based on the initial feature matrix $\mathbf{H}_0$ of the nodes as follows:
\begin{equation} \label{eq:initial}
\mathbf{Q}=\mathbf{H}_{0}\mathbf{W}_{\mathbf{Q}}, \mathbf{R}=\mathbf{H}_{0}\mathbf{W}_{\mathbf{R}},
\mathbf{\Delta}=\mathbf{H}_{0}\mathbf{W}_{\mathbf{\Delta}},
\end{equation}
where $\mathbf{W}_{\mathbf{Q}}$, $\mathbf{W}_{\mathbf{R}}$, and $\mathbf{W}_{\mathbf{\Delta}}$ are learnable parameters in the S3TL. 
This adaptive matrix generation enables the model to derive distinct $\bar{\mathbf{P}}$, $\bar{\mathbf{Q}}$, and $\mathbf{R}$ for each target node, allowing the model to adaptively weight the node’s features and its neighborhood information, thus enhancing its adaptability. 
The matrices $\bar{\mathbf{P}}$ and $\bar{\mathbf{Q}}$ are discretized as follows:
\begin{equation}
	\begin{aligned}
    \bar{\mathbf{P}}&={\rm exp}(\mathbf{\Delta} \mathbf{P}),\\
    \bar{\mathbf{Q}}&=(\mathbf{\Delta} \mathbf{P})^{-1}({\rm exp}(\mathbf{\Delta} \mathbf{P})-\mathbf{I})\cdot \mathbf{\Delta} \mathbf{Q},
\end{aligned}
\end{equation}
where ${\rm exp}(\cdot)$ refers to the exponential function with base $e$. 

It is worth noting that the state matrix $\mathbf{P}$ needs to be initialized in a special way named HiPPO-LegS~\cite{gu2020hippo}, which can enable $\mathbf{P}$ to select useful information for downstream tasks. 
The initialization process is defined as follows:
\begin{align} \label{eq:ini_P}
    \mathbf{P}[n,k]=-\begin{cases} 
(2n+1)^{1/2}(2k+1)^{1/2},  &\text{if }n>k\\
n+1, &\text{if } n=k\\
0, &\text{if } n<k
\end{cases}
\end{align}
where $n$ and $k$ are the indices along the two dimensions of the state matrix $\mathbf{P}$.

After alternating between the MAL and S3TL, the final node representation $\mathbf{Y}_l$ is derived by applying the output matrix $\mathbf{R}$ to the aggregated features from the target node’s $l$-order neighborhood:
\begin{equation}
	\begin{aligned} \label{eq:sql_1}
\mathbf{H}_{l}&=\bar{\mathbf{P}}\cdot\mathbf{H}_{l}+\bar{\mathbf{Q}}\cdot\mathbf{H}_{0}, \\ 
\mathbf{Y}_{l}&=\mathbf{R}\cdot\mathbf{H}_{l},
\end{aligned}
\end{equation}
where $\mathbf{Y}_l$ is the representation of the nodes after aggregating the information from its $l$-order neighborhood. 
$\mathbf{H}_{l}$ contains the compressed information from the $(l-1)$ hop neighborhood, which is adaptively refined through the discretized state matrix $\bar{\mathbf{P}}$ and the input matrix $\bar{\mathbf{Q}}$ to better adapt to downstream tasks.
This process enables the model to adaptively preserve important information from higher-order neighborhoods, discard redundant data, and retain the intrinsic features of the nodes.
Through this iterative process, the node features are refined and enriched, leading to more accurate node representations.

\subsection{Node State Prediction Layer (NSPL)}
While the MAL and S3TL effectively refine neighborhood aggregation, they are limited in distinguishing the significance of different nodes within the same neighborhood. 
This limitation becomes particularly critical when the model aggregates higher-order neighborhood information, as the number of nodes to be aggregated grows exponentially with the model depth, significantly increasing the presence of redundant nodes within the neighborhood.
To tackle this, we introduce the Node State Prediction Layer (NSPL). The main purpose of NSPL is further to regulate the information flow within the same-order neighborhood, allowing the model to prioritize the most relevant features and discard less important ones. This additional layer helps prevent the loss of key node-specific characteristics during message passing, ensuring that the model retains high-quality node representations even as it processes deeper graph layers.

In NSPL, we employ two parameter matrices $\mathbf{W}_1$ and $\mathbf{W}_2$ to predict the state of the target node while aggregating information from its neighborhood at various orders.
Specifically, the state vectors $\mathbf{S}^1_l$ and $\mathbf{S}^2_l$ are derived by applying the Gumbel-Softmax function to the node representations after neighborhood aggregation. This function allows for differentiable discrete sampling, which is essential for gradient-based optimization in deep learning. The state vectors $\mathbf{S}^1_l$ and $\mathbf{S}^2_l$ determine which neighborhood nodes contribute to the aggregation process, effectively controlling how much influence the information from each neighbor should have. The equations for generating the state vectors are as follows:
\begin{equation}
\begin{aligned}\label{eq:nspl_1}
    \mathbf{S}^{1}_{l}={\rm Gombel-Softmax}(\mathbf{Y}_{l-1}\mathbf{W}_{1}, \tau), \\ 
    \mathbf{S}^{2}_{l}={\rm Gombel-Softmax}(\mathbf{Y}_{l-1}\mathbf{W}_{2}, \tau),
\end{aligned}
\end{equation}
where $\mathbf{S}^1_l$ and $\mathbf{S}^2_l$ represent the predicted information flow when the target node aggregates data from its $l$-order neighborhood. The Gumbel-Softmax function is used for differentiable discrete sampling, enabling the model to optimize the information flow using gradient-based methods, even though the decision itself is discrete. The temperature parameter $\tau$ in the Gumbel-Softmax function controls the sharpness of the distribution. A lower value of $\tau$ results in more discrete outputs (closer to a one-hot vector), while a higher value introduce more randomness, promoting exploration during training. 
This mechanism allows for more flexible control of the information flow for each node during aggregation, enabling its contribution to be dynamically adjusted based on the scope of neighborhood aggregation.

To effectively manage the information flow between nodes and improve the model's adaptability, we introduce a crucial adjustment step based on the state vectors $\mathbf{S}^1_l$ and $\mathbf{S}^2_l$. These state vectors are used to modify the adjacency matrix $\mathbf{A}_l$ during subsequent message aggregation, enabling the model to control the information flow between different neighborhoods adaptively. This allows the model to fine-tune which neighbors' features should be aggregated, ensuring that only the most relevant information contributes to the node’s updated representation.
The modification of the adjacency matrix is formulated as follows:
\begin{align} \label{eq:modify_A}
    \mathbf{A}_{l}[i,j]=\begin{cases} 
1,  &\text{if }\mathbf{A}[i,j]=1\land \mathbf{S}^1_{l,j}=1\land \mathbf{S}^2_{l,i}=1\\
0, &\text{else}
\end{cases}
\end{align}

In this formula, $\mathbf{A}_l[i,j]$ determines whether there is a information flow between nodes $i$ and $j$ when aggregating the $l$-order neighborhood information. The decision is made based on the state vectors $\mathbf{S}^1_{l,j}$ and $\mathbf{S}^2_{l,i}$, which are learned from previous neighborhood aggregations. By regulating the information flow in this manner, the model ensures that only the most relevant features from each node's neighborhood are propagated, while irrelevant or redundant features are filtered out.

This dynamic adjustment process empowers the NSPL to adaptively control the information flow at each layer during aggregation. As a result, the model becomes more efficient in propagating meaningful features while avoiding the influence of noise or less important information. This targeted aggregation process improves the overall effectiveness of the graph representation learning, ensuring that each node’s updated representation is both accurate and informative.

\subsection{Total Complexity of \ourmethod}



The total time complexity of \ourmethod combines the complexities of MAL, S3TL, and NSPL. Since the alternating stacking of MAL and S3TL forms the core of the model, and the NSPL is applied after each layer, the total complexity for each layer is as follows: MAL and S3TL together contribute a time complexity of $\mathcal{O}(|\mathcal{E}|d + Nd^2)$ per layer, while NSPL adds an additional $\mathcal{O}(Nd^2)$ per layer. Assuming the model has $L$ layers, the total time complexity of \ourmethod is $\mathcal{O}(L \cdot |\mathcal{E}|d + L \cdot Nd^2)$. In practice, when the number of nodes and feature dimensions is large, the $L \cdot Nd^2$ term typically dominates. Thus, the overall complexity remains primarily influenced by the graph size and feature dimensions across all layers.
For a detailed implementation of \ourmethod, please refer to Algorithm~\ref{algor:mabgcn}.

\begin{algorithm}[h]
    \caption{Mamba-based Graph Convolution Network}\label{algor:mabgcn}

    \raggedright
    \textbf{Input}: Adjacency matrix $\mathbf{A} \in \mathbb{R}^{N \times N}$, feature matrix $\mathbf{X}\in \mathbb{R}^{N\times d}$, state matrix $\mathbf{P}$, learnable parameters $\mathbf{W}_{\mathbf{Q}}$, $\mathbf{W}_{\mathbf{R}}$, $\mathbf{W}_{\mathbf{\Delta}}$, $\mathbf{W}_{1}$, $\mathbf{W}_{2}$. \\
    \textbf{Output}: The updated node representations $\mathbf{Y}$.
    
    \begin{algorithmic}[1] 
    \STATE Compute $\mathbf{Q}$, $\mathbf{R}$ and $\mathbf{\Delta}$ via Eq.\ref{eq:initial};
    \WHILE{not convergent}
    \FOR{$l=1 \rightarrow L$}
    \STATE Compute $\hat{\mathbf{H}}_l$ via Eq.\ref{eq:mal};
    \STATE Compute $\mathbf{H}_l$ and $\mathbf{Y}_l$ via Eq.\ref{eq:sql_1};
    \STATE Compute $\mathbf{S}^1_l$ and $\mathbf{S}^2_l$ via Eq.\ref{eq:nspl_1};
    \STATE Modify the adjacency matrix $\mathbf{A}_l$ via Eq.\ref{eq:modify_A};
    \ENDFOR
    \STATE Obtain node representations $\mathbf{Y}$;
    \STATE Update all learnable parameters via back propagation;
    \ENDWHILE
    \RETURN Updated node representations $\mathbf{Y}$.
    \end{algorithmic}
\end{algorithm}

%% file: sections/experiment.tex
\section{Experiment}
In this section, we conduct a series of experiments to evaluate \ourmethod's performance, comparing it with several widely used GNN architectures. The aim is to assess \ourmethod's effectiveness as a new backbone for graph representation learning, focusing on its ability to address key challenges such as over-smoothing and adaptability to various graph structures. All experiments are performed on a system with an Intel(R) Xeon(R) Gold 5120 CPU and an NVIDIA L40 48G GPU.


\begin{table*}[t]
\renewcommand{\arraystretch}{1.3}
\centering
\scalebox{0.73}{
\begin{tabular}{c|ccc|cccc|c}
\toprule
\multicolumn{1}{c|}{\textbf{Datasets}}  & \multicolumn{1}{c}{\textbf{Cora}}& \multicolumn{1}{c}{\textbf{Citeseer}}& \multicolumn{1}{c}{\textbf{Pubmed}}& \multicolumn{1}{|c}{\textbf{Computers}}& \multicolumn{1}{c}{\textbf{Photo}}& \multicolumn{1}{c}{\textbf{Actor}}& \multicolumn{1}{c}{\textbf{Wisconsin}}& \multicolumn{1}{|c}{\textbf{Avg Rank}}\\
\midrule
\multirow{1}{*}{\textbf{GCN}}  &   87.04±0.70 (2) &   76.24±1.07 (2)  &  86.97±0.37 (2)  &   81.62±0.19 (2)    &    90.03±0.26 (2)      & 28.44±0.79 (2)& 53.75±3.25 (2)& 7.86\\
\multirow{1}{*}{\textbf{GAT}}
&  87.65±0.24 (2)    &   76.20±0.27 (2)  &   87.39±0.11 (2) &  82.76±0.75 (2) &   90.25±0.92 (2) & 29.92±0.23 (2) &55.49±3.14 (2)& 6.57\\
\multirow{1}{*}{\textbf{SGC}}
&   86.96±0.87 (2)  &   75.82±1.06 (2)  &    87.36±0.29 (2)&    84.13±0.87 (2)&  92.34±0.38 (2) &  26.73±1.04 (2)&  50.39±2.94 (2)& 8.00\\ \midrule
                \multirow{1}{*}{\textbf{APPNP}}
&  87.71±0.76 (4)    &   76.66±1.22 (2) &  87.76±0.43 (10)  &   84.51±0.34 (4)   &    88.97±0.96 (6)     & 29.68±0.72  (2)  & 59.80±1.96 (2)& 5.71\\
                \multirow{1}{*}{\textbf{GCNII}}
&   \cellcolor{grey!80}88.07±0.93 (64)  &  \cellcolor{grey!80}77.99±1.01 (64) &  \cellcolor{grey!80}90.15±0.31 (64) &    84.71±0.40 (8) &  92.46±0.70 (4)   &  37.31±0.55 (8) & 80.19±6.29 (10)&  2.86\\
\multirow{1}{*}{\textbf{GPRGNN}}
&   87.75±0.62 (6)   &   \underline{77.08±1.06} (4)  &  \underline{89.36±0.33} (6)   &  87.43±0.49 (8)  &  \underline{94.36±0.31} (10)& 33.87±0.58 (8)&  81.02±2.94 (4)& \underline{2.71}\\ 
\multirow{1}{*}{\textbf{SSGC}}&   87.40±0.87 (6)&   75.80±1.03 (2) &  87.67±0.38 (4) &  85.95±0.78 (4)  & 93.39±0.33 (4)&  29.15±0.69 (2)&  52.75±1.76  (2)& 6.43\\ 
\multirow{1}{*}{\textbf{GGCN}}& 87.73±1.24 (4) &  76.63±1.49 (8) &  89.08±0.47 (5) &  \underline{90.36±0.52} (2) & 94.23±0.65 (6)& \underline{37.54±1.46} (8)& \underline{85.88±4.19} (4)& 3.14\\
\midrule
\multirow{1}{*}{\textbf{\ourmethod (ours)}}
&   \underline{87.79±0.60} (10)   &   76.68±0.96 (6)  &  89.32±0.24 (8)   &  \cellcolor{grey!80}90.39±0.21 (4) &   \cellcolor{grey!80}94.41±0.75 (2) & \cellcolor{grey!80}37.97±0.91 (10)&   \cellcolor{grey!80}86.27±2.16 (8)& \cellcolor{grey!80}1.71\\
\bottomrule
\end{tabular}}
\caption{Summary of classification accuracy (\%) results. The best result for each benchmark is highlighted with a gray background, and the second-best result is emphasized with an underline. The layer configurations that achieve the best performance are recorded in brackets.} \label{tab:performance_com}
\end{table*}

\subsection{Experimental Settings}
\textbf{Datasets: }We evaluate our method on a variety of datasets across different domains, focusing on full-supervised node classification tasks. The datasets include three citation graph datasets (\textbf{Cora, Citeseer, Pubmed}), two web graph datasets (\textbf{Computers, Photo}), and two heterogeneous graph datasets (\textbf{Actor, Wisconsin}).
For citation and heterogeneous graph datasets, we use the feature vectors, class labels, and 10 random splits as proposed in~\cite{chen2020simple}. For the web graph datasets, the same components are used following the protocol in~\cite{he2021bernnet}. Detailed statistics and descriptions of these datasets can be found in Appendix A.1.

\begin{figure*}[h!]
\centering
\subfigure{\includegraphics[width=0.24\linewidth]{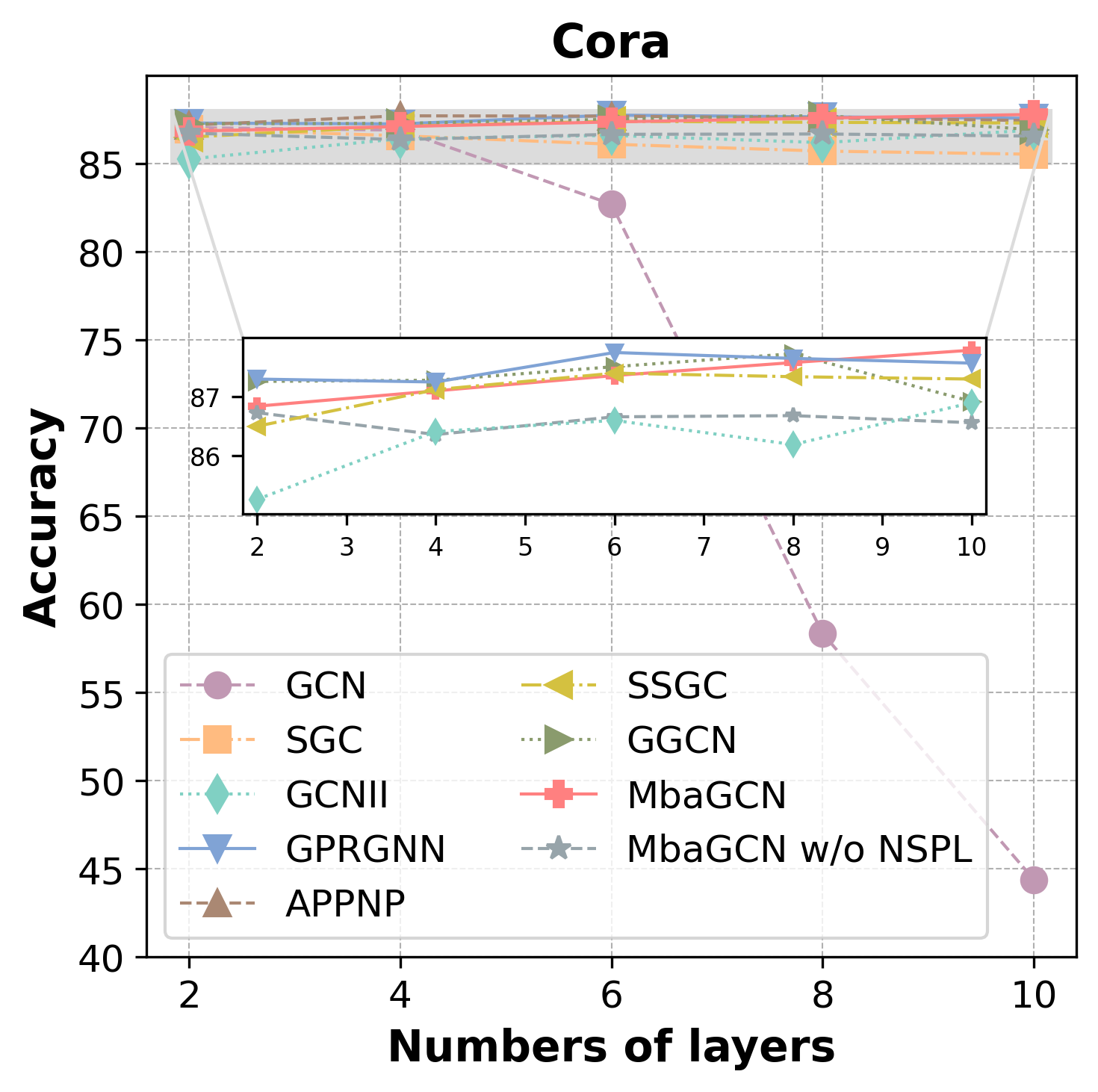}}%
\subfigure{\includegraphics[width=0.24\linewidth]{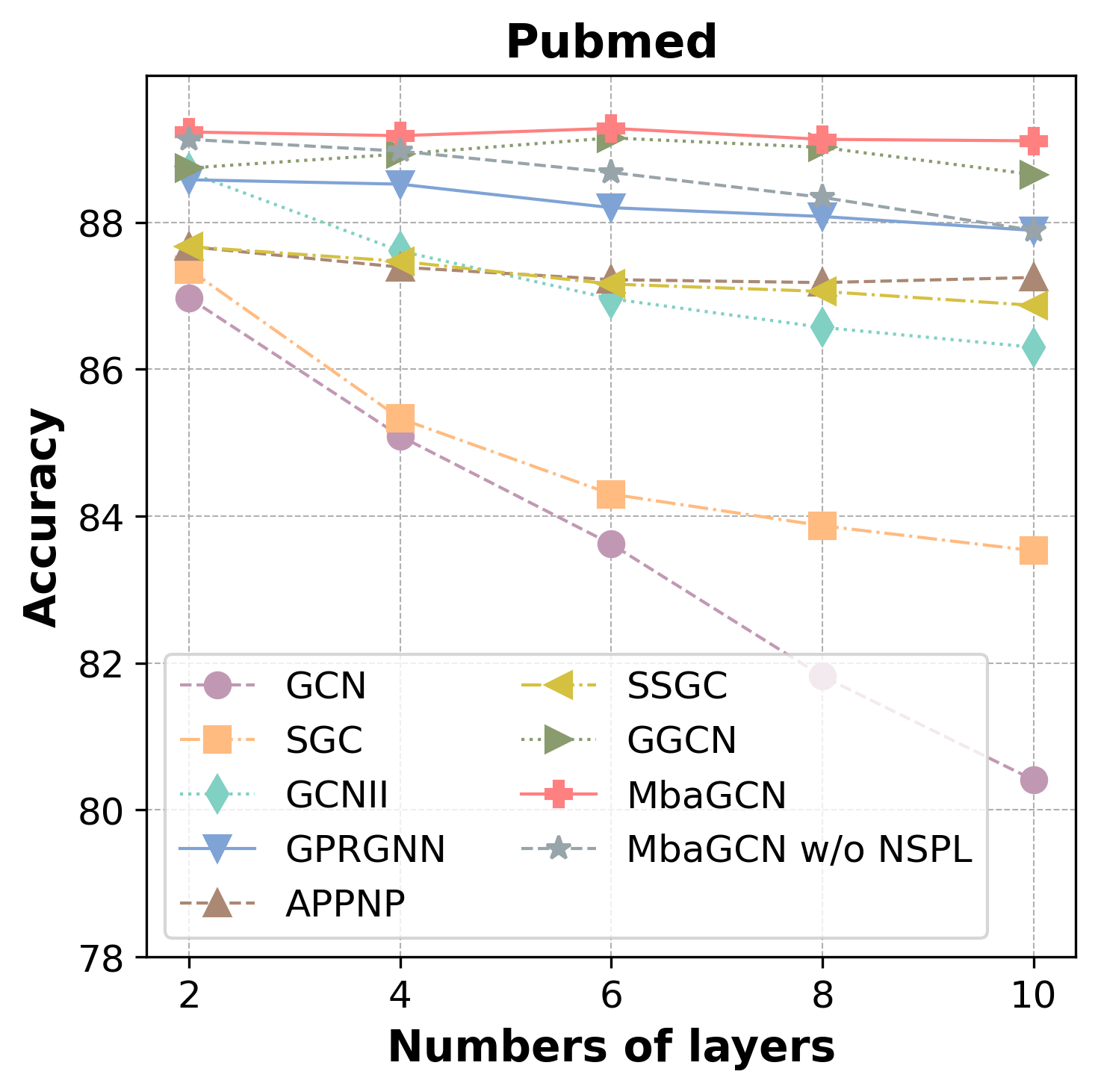}}%
\subfigure{\includegraphics[width=0.24\linewidth]{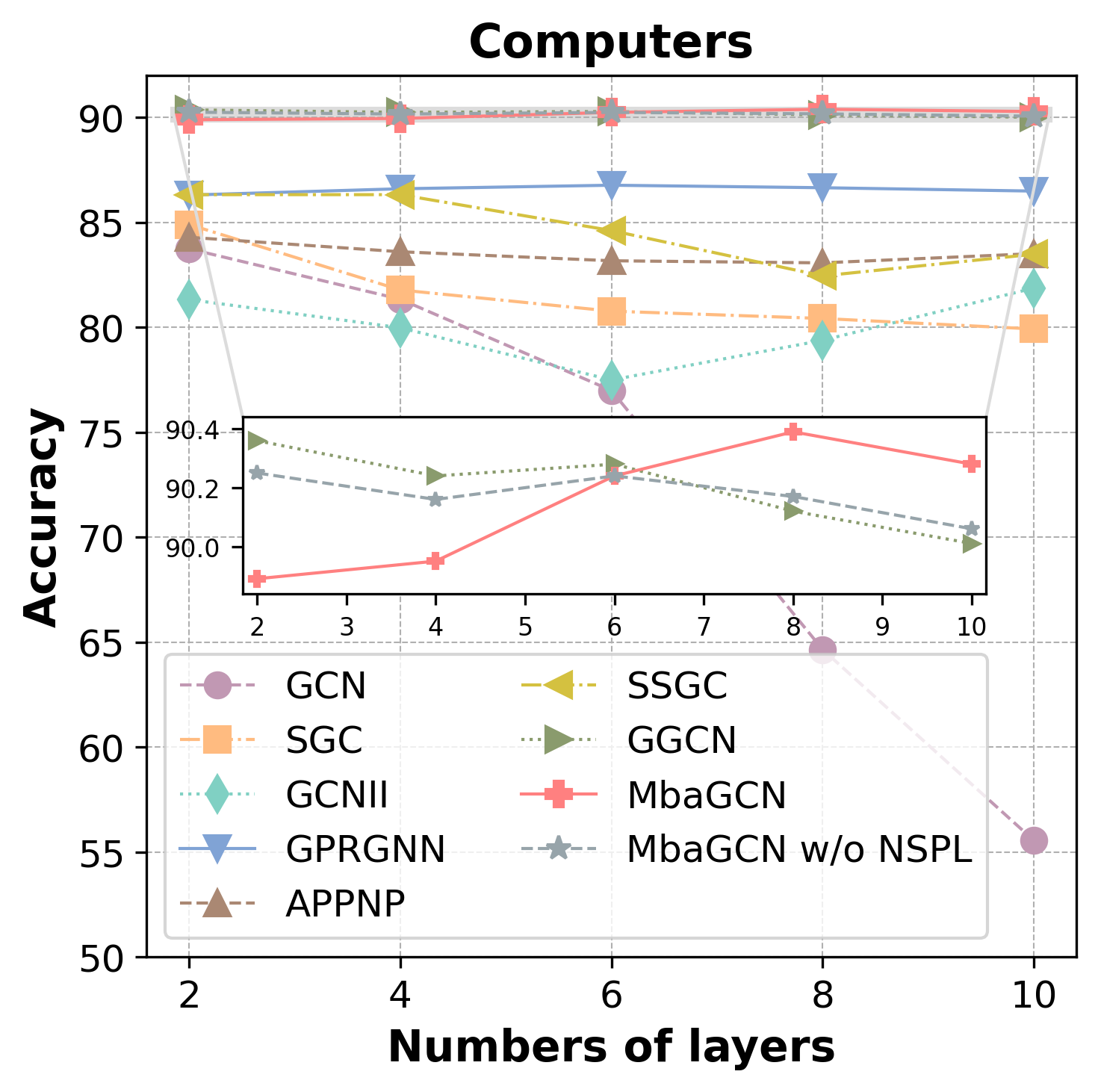}}%
\subfigure{\includegraphics[width=0.24\linewidth]{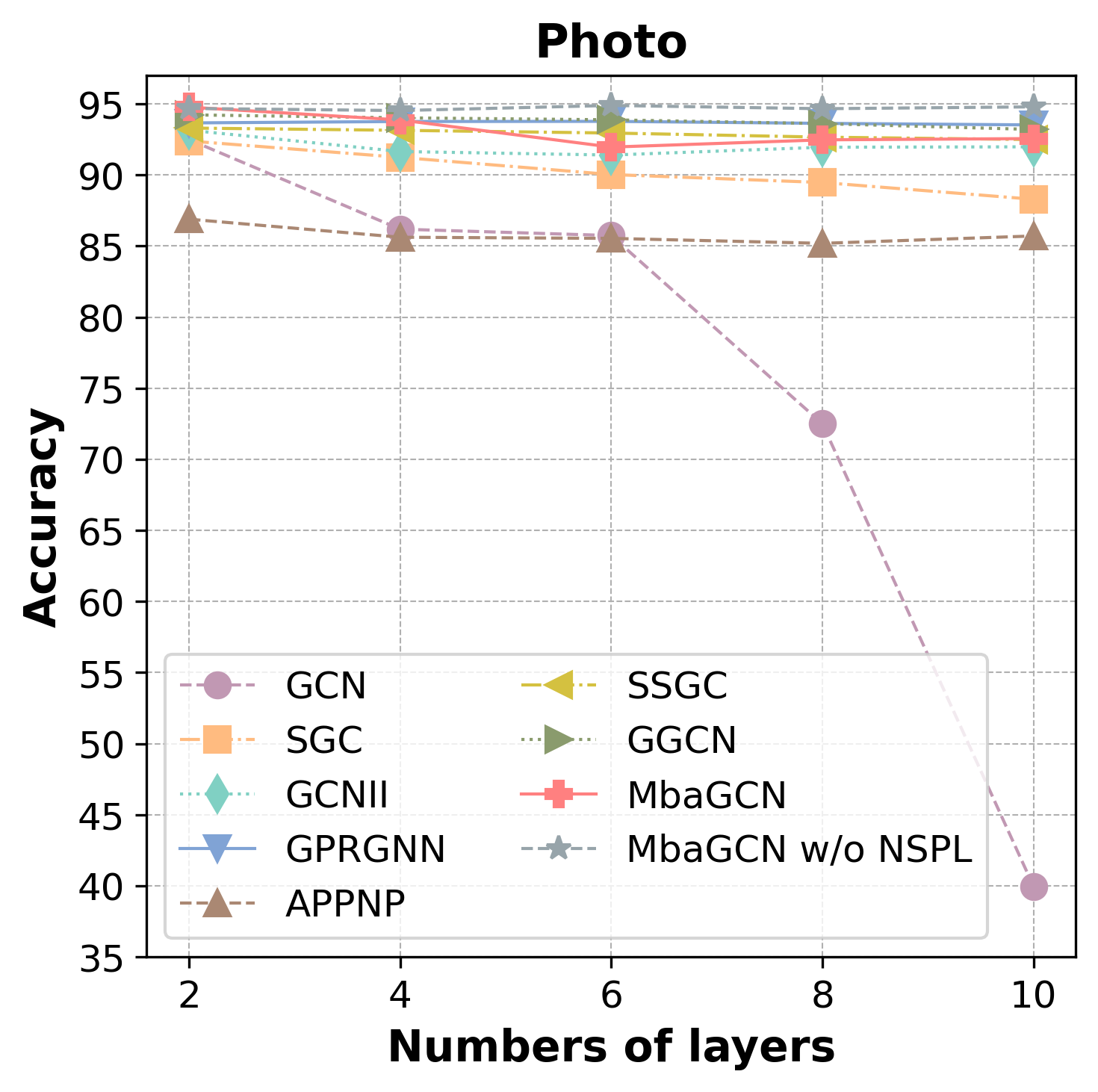}}%
\caption{Performance of baselines and the proposed \ourmethod with 2/4/6/8/10 layers.}
\label{fig:ablation_study}
\end{figure*}

\noindent\textbf{Baselines: }To evaluate the effectiveness of \ourmethod, we compare it with several representative GNN models, including classical models like \textbf{GCN}~\cite{kipf2016semi}, \textbf{GAT}~\cite{velivckovic2017graph}, and \textbf{SGC}~\cite{wu2019simplifying}, as well as deep GNN models such as \textbf{APPNP}~\cite{gasteiger2018predict}, \textbf{GCNII}~\cite{chen2020simple}, \textbf{GPRGNN}~\cite{chien2020adaptive}, \textbf{SSGC}~\cite{zhu2021simple}, and \textbf{GGCN}~\cite{yan2022two}. Further details about these baseline models can be found in Appendix A.2.

\subsection{Performance Evaluation of \ourmethod}
\textbf{Q: Does \ourmethod outperform baseline models across various datasets?} Yes, \ourmethod consistently achieves the highest average rank across all datasets, demonstrating its overall adaptability and robustness.

\noindent$\rhd$ \textsf{Performance across Diverse Datasets:} As shown in Tab.\ref{tab:performance_com}, \ourmethod consistently outperforms all baseline models, achieving an average rank of 1.71, and ranks first on six out of eight datasets. This demonstrates its superior adaptability across both homophilic and heterophilic graph structures. On citation graph datasets like Cora, Citeseer, and Pubmed, \ourmethod achieves competitive results, closely matching or surpassing the top-performing models. For example, on Cora, \ourmethod achieves an accuracy of 87.79\%, just marginally lower than GCNII. This shows its effectiveness in handling strongly homophilic graphs and datasets with more complex structures. The model's strong performance can be attributed to its ability to flexibly aggregate neighborhood information, which allows it to capture local and global features while avoiding over-smoothing effectively, a common issue in deeper models.

\noindent$\rhd$ \textsf{Superior Performance on Heterophilic Datasets:} \ourmethod excels on heterophilic datasets, where traditional GNNs struggle due to adjacent nodes having dissimilar features. Standard GNNs often aggregate neighborhood information indiscriminately, leading to ineffective learning. In contrast, \ourmethod utilizes a adaptive aggregation mechanism inspired by Mamba, which adapts the aggregation process based on the relevance of the information. This allows \ourmethod to retain meaningful features and discard irrelevant ones, especially in heterophilic graphs.
On the Wisconsin dataset, a typical heterophilic graph, \ourmethod achieves an accuracy of 86.27\%, outperforming all other models. Similarly, on the Actor dataset, \ourmethod achieves 37.97\%, demonstrating its robustness in heterophilic scenarios where traditional GNNs tend to underperform.

\noindent$\rhd$ \textsf{Consistency and Robustness:} In addition to excelling in heterophilic settings, \ourmethod maintains high performance across a wide variety of graph structures. While models like GCNII, GPRGNN, and GGCN perform well on specific datasets (e.g., GCNII on homophilic graphs), their overall rankings are lower compared to \ourmethod, highlighting the latter’s more consistent and robust performance across multiple types of datasets. This suggests that the flexibility and adaptability of the selective aggregation approach in \ourmethod allow it to handle a broad range of graph complexities and maintain high accuracy.

\begin{table*}[t]
\renewcommand{\arraystretch}{1.3}
\centering
\scalebox{0.73}{
\begin{tabular}{c|ccccc|ccccc}
\toprule
\multicolumn{1}{c|}{\textbf{Layers}}  & \multicolumn{1}{c}{\textbf{2}}& \multicolumn{1}{c}{\textbf{4}}& \multicolumn{1}{c}{\textbf{6}}& \multicolumn{1}{c}{\textbf{8}}& \multicolumn{1}{c}{\textbf{10}}& \multicolumn{1}{|c}{\textbf{2}}& \multicolumn{1}{c}{\textbf{4}}& \multicolumn{1}{c}{\textbf{6}}& \multicolumn{1}{c}{\textbf{8}}& \multicolumn{1}{c}{\textbf{10}}\\
\midrule

\textbf{Dataset}&\multicolumn{5}{c|}{\textbf{Actor}}&\multicolumn{5}{c}{\textbf{Wisconsin}} \\
\cmidrule{1-11}
\multirow{1}{*}{\textbf{GCN}}  &  28.44±0.79
 & 27.18±0.51 & 26.93±0.55 & 26.56±0.38 & 26.55±0.43& 53.75±3.25 & 51.96±2.95 & 48.63±3.53 & 48.04±4.12& 47.00±4.51\\
\multirow{1}{*}{\textbf{SGC}}& 26.73±1.04  & 24.98±0.47 & 24.97±0.71 & 25.09±0.76 & 25.19±0.79 & 
 50.39±2.94 & 50.00±4.12 & 50.20±3.33 & 49.41±3.53 & 48.82±3.92\\ \midrule
                \multirow{1}{*}{\textbf{APPNP}}
& 29.68±0.72  & 28.77±0.75 & 28.38±0.71 & 28.55±0.82 & 28.35±0.66 & 59.80±1.96 & 59.22±2.36 & 58.63±2.35 & 59.41±3.14 & 57.84±2.75\\
                \multirow{1}{*}{\textbf{GCNII}}
& 36.31±0.55  & 36.37±0.76 & 37.12±0.57 & 37.31±0.60 & 36.91±0.51 & 79.25±2.75 & 79.67±3.14 & 79.75±2.55 & 79.85±2.75 & 80.19±2.75\\
\multirow{1}{*}{\textbf{GPRGNN}}
& 32.62±0.66  & 32.62±0.97 & 33.34±0.60 & 33.87±0.58 &  33.60±0.58& 79.08±3.92 & 81.02±2.94&  78.12±2.95& 76.67±2.16 & 75.10±2.94\\ 
\multirow{1}{*}{\textbf{SSGC}}& 29.15±0.69  & 28.51±0.79 & 28.55±0.72 & 28.56±0.83 & 28.64±1.00 & 52.75±1.76& 52.75±2.94 & 49.61±2.55 & 50.20±3.33 & 52.75±2.75\\ 
\multirow{1}{*}{\textbf{GGCN}}& 37.22±1.29  & \cellcolor{grey!80}37.46±1.16& \cellcolor{grey!80}37.50±1.42 & \underline{37.54±1.46} & \underline{37.25±1.28} & 84.51±4.06& \underline{85.88±4.19} & 84.12±4.51 & 84.31±4.38&84.12±4.76\\ \cmidrule{1-11}
\multirow{1}{*}{\textbf{\ourmethod (ours)}}& \cellcolor{grey!80}37.47±0.76  & 37.10±0.70
& \underline{37.42±0.72} & \cellcolor{grey!80}37.65±0.72 &\cellcolor{grey!80}37.97±0.91& \underline{85.29±2.35}& \cellcolor{grey!80}85.88±1.57 & \cellcolor{grey!80}85.49±1.96 & \cellcolor{grey!80}86.27±2.16 & \cellcolor{grey!80}85.49±3.14\\
\multirow{1}{*}{\textbf{\ourmethod w/o NSPL}}& \underline{37.43±0.61}  & \underline{37.32±0.83}
& 37.33±0.92 & 37.41±0.48 &37.31±0.76& \cellcolor{grey!80}85.31±3.37& 85.28±2.77 & \underline{85.32±2.39} & \underline{85.27±1.68} & \underline{85.30±2.84}\\
\bottomrule
\end{tabular}}
\caption{Classification accuracy (\%) comparison under different layer configurations. The best result under the same layer configuration is highlighted with a gray background, and the second-best result is emphasized with an underline.} \label{tab:performance_com_under_dif_lay}
\end{table*}

\subsection{Impact of Layer Depth on Performance}
\textbf{Q: How does \ourmethod perform under different layer depths compared to baseline models?}
\ourmethod consistently maintains high performance and stability across varying numbers of layers, demonstrating its robustness in deeper architectures and its ability to mitigate over-smoothing.

\noindent$\rhd$ \textsf{Performance Trends Across Layer Depths:}
Tab.\ref{tab:performance_com_under_dif_lay} summarizes the classification accuracy (\%) of various GNN models across 2, 4, 6, 8, and 10 layers on the Actor and Wisconsin datasets. The results reveal a clear trend: GCN experiences significant performance degradation beyond 2 layers, highlighting its susceptibility to over-smoothing. Other deep GNN methods (e.g., GCNII) show better resilience, but their performance often peaks with shallow architectures, declining slightly at deeper depths. For example, APPNP on Actor and GCNII on Cora demonstrate marginal declines after 4 layers.
The experimental results for the other two datasets (Cora and Citeseer) can be found in Appendix A.3.

\noindent$\rhd$ \textsf{\ourmethod’s Stability at Greater Depths:}
In contrast, \ourmethod maintains consistent and competitive performance across all tested layer depths. Its selective state space mechanism enables it to prioritize important features and avoid the indiscriminate propagation of redundant information. For instance, on the Wisconsin dataset, \ourmethod achieves top accuracy even at 10 layers, significantly outperforming other models that degrade at similar depths. On homophilic datasets like Cora and Pubmed (see Fig.~\ref{fig:ablation_study}), \ourmethod achieves accuracy levels comparable to or exceeding the best-performing models, demonstrating its adaptability to both homophilic and heterophilic graph structures.

\begin{table}[t]
\renewcommand{\arraystretch}{1.2}
\centering
\scalebox{0.9}{
\begin{tabular}{c|ccc}
\toprule
\multicolumn{1}{c|}{\textbf{Datasets}} & \multicolumn{1}{c}{\textbf{Citeseer}}& \multicolumn{1}{c}{\textbf{Actor}}&\multicolumn{1}{c}{\textbf{Wisconsin}}  \\
\multirow{1}{*}{\textbf{Best}} &  6 & 10 &   8\\ \midrule
\multirow{1}{*}{\textbf{MbaGCN}} &   \textbf{76.68±0.96} & \textbf{37.97±0.91} &   \textbf{86.27±2.16}\\ \midrule
    \multirow{1}{*}{\textbf{MbaGCN w/o HL}}
&   75.87±0.73   & 35.39±2.49 &  82.55±2.35\\
\multirow{1}{*}{\textbf{Decline}} &  1.06\%  &  6.79\% &  4.31\%  \\ \midrule
\multirow{1}{*}{\textbf{MbaGCN w/o IR}}
&   74.35±0.89   & 34.26±0.68 & 81.96±1.18 \\
\multirow{1}{*}{\textbf{Decline}} &  3.04\%  & 9.77\% & 5.00\%  \\
\bottomrule
\end{tabular}}
\caption{Ablation experiments of HL~(HiPPO-LegS), IR~(Input-Related) in proposed selective state space transition layer (S3TL).} \label{tab:performance_abl}
\end{table}

\subsection{Ablation Study}
\textbf{Q: How do NSPL, HiPPO-LegS (HL), and Input-Related (IR) contribute to the performance of \ourmethod, particularly in deeper layers?} These components collectively enhance \ourmethod’s adaptability, mitigate over-smoothing, and preserve feature distinctiveness in deeper layers.

\noindent$\rhd$ \textsf{Impact of NSPL on Performance Across Depths:}
As shown in Fig.\ref{fig:ablation_study} and Tab.\ref{tab:performance_com_under_dif_lay}, NSPL significantly enhances \ourmethod’s ability to maintain performance in deeper architectures. Without NSPL, \ourmethod performs best at shallow depths (e.g., 2 layers) but experiences a sharp decline as the depth increases. This is due to the model's inability to regulate message flow within same-order neighborhoods, resulting in excessive information propagation or dilution of higher-order features. In contrast, incorporating NSPL allows dynamic control of message flow, preserving critical features and improving feature aggregation. For example, on the Wisconsin dataset, \ourmethod with NSPL achieves peak performance at 8 layers, while the ablated version struggles. However, on dense datasets like Photo, NSPL's impact diminishes, likely due to optimization difficulties in dense graph structures, highlighting a potential area for future improvement.

\noindent$\rhd$ \textsf{Ensuring Robust Feature Propagation with HiPPO-LegS (HL):}
HL in Eq.\ref{eq:initial} plays a pivotal role in maintaining robust feature propagation across deeper architectures. By ensuring efficient state transitions, HL minimizes the risk of feature degradation that often arises in deep GNNs due to over-smoothing. As shown in Tab.\ref{tab:performance_abl}, removing HL results in significant performance declines, such as a 6.79\% drop on the Actor dataset and a 2.48\% drop on Wisconsin. These declines highlight HL's critical role in preserving distinct node representations while enabling effective information aggregation across layers. Furthermore, HL's impact becomes increasingly pronounced as the network depth grows, showcasing its ability to adapt state transitions dynamically and mitigate the compounding effects of over-smoothing.

\noindent$\rhd$ \textsf{Dynamic Adaptability Through Input-Related (IR) Matrices:}
IR in Eq.\ref{eq:ini_P} enhances the adaptability of \ourmethod by generating state matrices that dynamically reflect node- and neighborhood-specific characteristics. This flexibility allows \ourmethod to balance the influence of local and global information, ensuring that critical features are neither overshadowed by higher-order information nor lost in the aggregation process. As depicted in Tab.\ref{tab:performance_abl}, the absence of IR results in notable performance degradation, such as a 9.77\% accuracy drop on Actor and a 4.12\% drop on Wisconsin. Compared to HL, IR demonstrates an even greater influence on model performance, particularly in datasets with diverse graph structures. 
This highlights IR’s key role in adapting state transitions to each graph’s structure, ensuring consistent performance across different depths and complexities.

%% file: sections/conclution.tex
\section{Conclution}
This paper introduces \textbf{MbaGCN}, a novel architecture that integrates the Mamba paradigm into GNNs, addressing key challenges such as the loss of node-specific features in deeper architectures. By alternating between Message Aggregation Layers (MAL) and Selective State Space Transition Layers (S3TL), and incorporating the Node State Prediction Layer (NSPL), \ourmethod enables adaptive aggregation and propagation of information.
Experimental results demonstrate \ourmethod's strong performance across diverse datasets, particularly on heterophilic graphs. Ablation studies highlight the importance of key components like HiPPO-LegS (HL) and Input-Related (IR) in improving model adaptability and mitigating over-smoothing.
While promising, \ourmethod faces challenges in dense graphs, suggesting opportunities for future work to optimize its components and extend its applicability to dynamic and multi-modal graphs. This study establishes a foundation for adaptive and scalable GNN architectures inspired by the Mamba paradigm.

%% file: sections/acknowledgment.tex
\section*{Acknowledgments}
This work was supported by a grant from the National Natural Science Foundation of China under grants (No.62372211,  62272191), and the Science and Technology Development Program of Jilin Province (No.20250102216JC).

%% file: sections/contributionstatement.tex
\section*{Contribution Statement}
Xin He and Yili Wang designed the experiments. Xin He performed the experiments. Yili Wang and Wenqi Fan analyzed the data. All authors contributed to writing and reviewing the manuscript. Xin He and Yili Wang contributed equally to this work. Xin Wang is the corresponding author.

%% file: sections/appendix.tex
\appendix

\section{Experiments Detail}


\begin{table}[hb]
\renewcommand{\arraystretch}{1.2}
\centering
\scalebox{1}{
\begin{tabular}{c|cccc}
\toprule
\multicolumn{1}{c|}{\textbf{Datasets}}  & \multicolumn{1}{c}{\textbf{Nodes}}& \multicolumn{1}{c}{\textbf{Edges}}& \multicolumn{1}{c}{\textbf{Features}}& \multicolumn{1}{c}{\textbf{Classes}}\\
\midrule
 \multicolumn{1}{c|}{\textbf{Cora}} &  2,708  & 10,556  &  1,433  &   7  \\
\multicolumn{1}{c|}{\textbf{Citeseer}}& 3,327   &   9,104&  3,703  &  6   \\
\multicolumn{1}{c|}{\textbf{Pubmed}}& 19,717   &   88,648&  500  &  3   \\
\multicolumn{1}{c|}{\textbf{Computers}}& 23,752   &  491,722 &   767 &  10   \\
\multicolumn{1}{c|}{\textbf{Photo}}& 7,650   &  238,162 &   745 &  8   \\
\multicolumn{1}{c|}{\textbf{Actor}}& 7,600   &  30,019 &  932 & 5   \\
\multicolumn{1}{c|}{\textbf{Wisconsin}}& 251  & 515 &  1,703 &  5   \\
\bottomrule
\end{tabular}}
\caption{Dataset statistics.} \label{tab:datasets}
\end{table}

    \begin{table*}[]
\renewcommand{\arraystretch}{1.3}
\centering
\scalebox{0.75}{
\begin{tabular}{c|ccccc|ccccc}
\toprule
\multicolumn{1}{c|}{\textbf{Layers}}  & \multicolumn{1}{c}{\textbf{2}}& \multicolumn{1}{c}{\textbf{4}}& \multicolumn{1}{c}{\textbf{6}}& \multicolumn{1}{c}{\textbf{8}}& \multicolumn{1}{c}{\textbf{10}}& \multicolumn{1}{|c}{\textbf{2}}& \multicolumn{1}{c}{\textbf{4}}& \multicolumn{1}{c}{\textbf{6}}& \multicolumn{1}{c}{\textbf{8}}& \multicolumn{1}{c}{\textbf{10}}\\
\midrule
\textbf{Dataset}&\multicolumn{5}{c|}{\textbf{Cora}}&\multicolumn{5}{c}{\textbf{Citeseer}} \\
\cmidrule{1-11}
\multirow{1}{*}{\textbf{GCN}}  & 87.04±0.70  & 86.90±0.78 & 82.70±0.78 & 58.35±5.05 & 44.39±2.48 & 76.24±1.07 & 75.59±1.00 & 66.86±2.66 & 41.17±4.50& 37.42±3.15\\
\multirow{1}{*}{\textbf{SGC}}& 86.96±0.87  & 86.58±0.74 & 86.10±0.60 & 85.71±0.74 & 85.53±0.78 & 75.82±1.06& 75.15±0.72 & 74.46±0.90 & 74.04±0.90 & 73.87±0.89\\ \midrule
                \multirow{1}{*}{\textbf{APPNP}}
& 87.18±0.89  & 87.71±0.76 & \underline{87.69±0.91} & 87.59±0.83 & 87.44±0.91 & \underline{76.66±1.22} & 75.64±1.04 & 75.96±0.95 &75.97±0.92 & 75.80±1.09\\
                \multirow{1}{*}{\textbf{GCNII}}
& 85.26±0.76  & 86.41±0.91 & 86.60±1.39 & 86.19±1.21 & 86.90±0.87 & 74.72±1.24 & 74.48±1.13 & 74.94±1.36 & 75.63±1.18 & 76.01±1.26\\
\multirow{1}{*}{\textbf{GPRGNN}}
& \cellcolor{grey!80}87.30±0.87  & \underline{87.25±0.79} & \cellcolor{grey!80}87.75±0.62 & \underline{87.65±0.91}
& \underline{87.57±0.82} & \cellcolor{grey!80}76.89±1.28 & \cellcolor{grey!80}77.08±1.06 &  \cellcolor{grey!80}76.96±1.01& \underline{76.54±1.00}
& \underline{76.57±0.91}\\ 
\multirow{1}{*}{\textbf{SSGC}}& 86.50±0.72  & 87.12±0.87 & 87.40±0.87 & 87.34±0.78 & 87.30±0.87 & 75.80±1.03& 75.69±1.23 & 75.55±0.94 & 75.57±0.88 & 75.21±0.76\\ 
\multirow{1}{*}{\textbf{GGCN}}&  \underline{87.26±1.27} & \cellcolor{grey!80}87.28±1.28 & 87.51±1.42 & \cellcolor{grey!80}87.73±1.24 & 86.92±0.99 & 76.63±1.54& \underline{76.53±1.43} & 76.47±1.80 & \cellcolor{grey!80}76.63±1.49 & \cellcolor{grey!80}76.60±1.60\\ \cmidrule{1-11}
\multirow{1}{*}{\textbf{MbaGC (ours)}}& 86.84±0.85  & 87.10±0.80 & 87.36±0.76 & 87.58±1.29 & \cellcolor{grey!80}87.79±0.60 & 76.03±0.91 & 75.91±1.20 & \underline{76.68±0.96} & 76.03±0.96& 75.80±1.40\\
\multirow{1}{*}{\textbf{MbaGCN w/o NSPL}}& 86.73±0.75&86.36±0.91&86.66±0.86&86.68±1.18&86.56±0.74& 76.23±0.94& 76.28±1.49 & 76.16±0.88 & 76.19±1.41 & 76.12±1.22\\
\bottomrule
\end{tabular}}
\caption{Classification accuracy (\%) comparison under different layer configurations. The best result under the same layer configuration is highlighted with a gray background, and the second-best result is emphasized with an underline.} \label{tab:performance_com_under_dif_lay_app}
\end{table*}

\subsection{Dataset}\label{appendix:dataset}
The statistics are listed in Tab.~\ref{tab:datasets}.
Cora, Citeseer, Pubmed are citation graph datasets, Computers and Photo are web graph datasets, Actor and Wisconsin are heterogeneous graph datasets.
We utilize the feature vectors, class labels, and 10 random splits as proposed by~\cite{chen2020simple} for citation graph datasets and heterogeneous graph datasets. 
We use the feature vectors, class labels, and 10 random splits as detailed in~\cite{he2021bernnet} for web graph datasets.
\begin{itemize}
    \item \textbf{Cora \& Citeseer \& Pubmed}: These three datasets are benchmark citation network datasets, where nodes represent papers and edges denote citation relationships.
    \item \textbf{Computers \& Photo}: These two datasets are commonly used node classification datasets from the Amazon co-purchase graph. In these datasets, nodes represent goods, and edges indicate that two goods are frequently bought together. 
    \item \textbf{Actor}: This is an actor-only subgraph of the film-director-actor-writer network, where nodes represent actors, and edges denote co-occurrence on the same Wikipedia page. Node features are derived from keywords in the Wikipedia pages.
    \item \textbf{Wisconsin}: This dataset consists of academic papers or webpages, with each paper or webpage treated as a node. Citation or hyperlink relationships between them are represented as edges.
\end{itemize}

\subsection{Baselines}\label{appendix:baseline}
\begin{itemize}
    \item \textbf{GCN}: This method is the most classic graph neural network, which enhances the model's stability and scalability through the first-order approximation.
    \item \textbf{GAT}: This method is a graph neural network that uses an attention mechanism to explore node attributes across the graph, allowing for implicit weighting of different nodes within a neighborhood.
    \item \textbf{SGC}: This method is a simplified variant of GCN that removes nonlinearities and learnable parameter matrices between graph convolution layers.
    \item \textbf{APPNP}: This method improves the propagation scheme of GCN by leveraging personalized PageRank to enhance the performance of GCN models.
    \item \textbf{GCNII}: This method is a variant of GCN that incorporates residual connections and identity mapping, effectively alleviating the over-smoothing phenomenon.
    \item \textbf{GPRGNN}: This method is a GNN model that adapts the PageRank algorithm within a GNN to capture node importance, and improve performance on tasks like classification and link prediction, while also alleviating the over-smoothing issue.
    \item \textbf{SSGC}: This method mitigates over-smoothing by using a modified Markov diffusion kernel, which symmetrically scales the aggregation process to preserve node feature distinctiveness in deeper layers.
    \item \textbf{GGCN}: This method solves over-smoothing in both homophilic and heterophilic graphs through structure-based and feature-based edge correction.
\end{itemize}

\subsection{Impact of Layer Depth on Performance}
Tab.~\ref{tab:performance_com_under_dif_lay_app} summarizes the classification accuracy (\%) of various GNN models across 2, 4, 6, 8, and 10 layers on the Cora and Citeseer datasets.
GCN suffers from significant performance degradation beyond 2 layers due to its susceptibility to over-smoothing. In contrast, deeper GNN methods, such as GCNII, exhibit greater robustness, with their performance remaining stable or even improving steadily as the number of layers increases.

From the performance changes observed in our proposed MbaGCN and its ablation version MbaGCN w/o NSPL as the number of layers increases, the introduction of NSPL enhances the model's ability to mitigate the over-smoothing problem. However, it also reduces the stability of the model's performance. For instance, on the Citeseer dataset, the inclusion of NSPL results in greater fluctuations in MbaGCN's performance as the number of layers increases, which may be attributed to the increased optimization difficulty of NSPL in deeper architectures.

\subsection{Time And Memory Cost}

\begin{figure}[h]
\centering
{\subfigure
{\includegraphics[width=0.49\linewidth]{{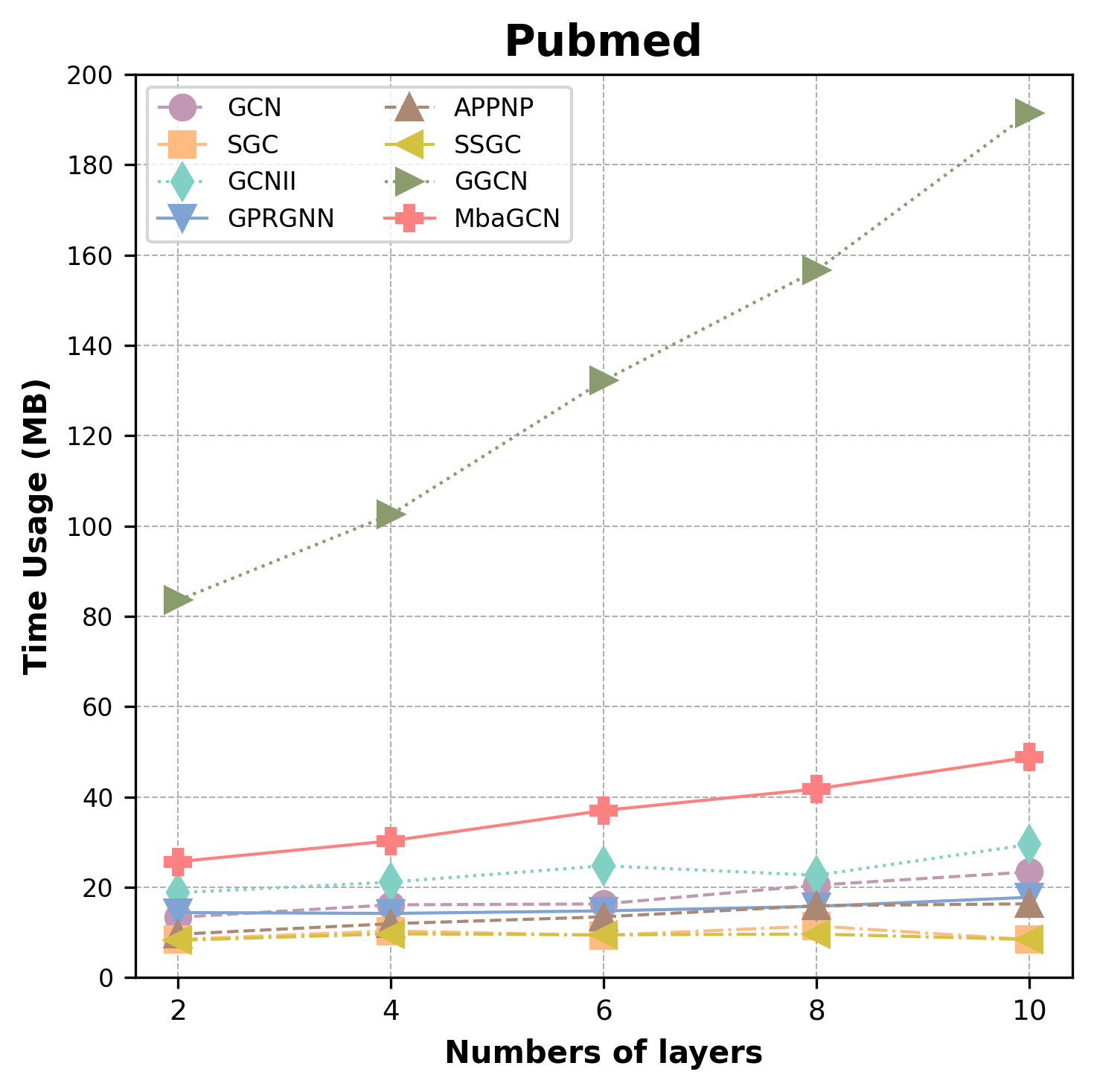}}}}%
{\subfigure
{\includegraphics[width=0.49\linewidth]{{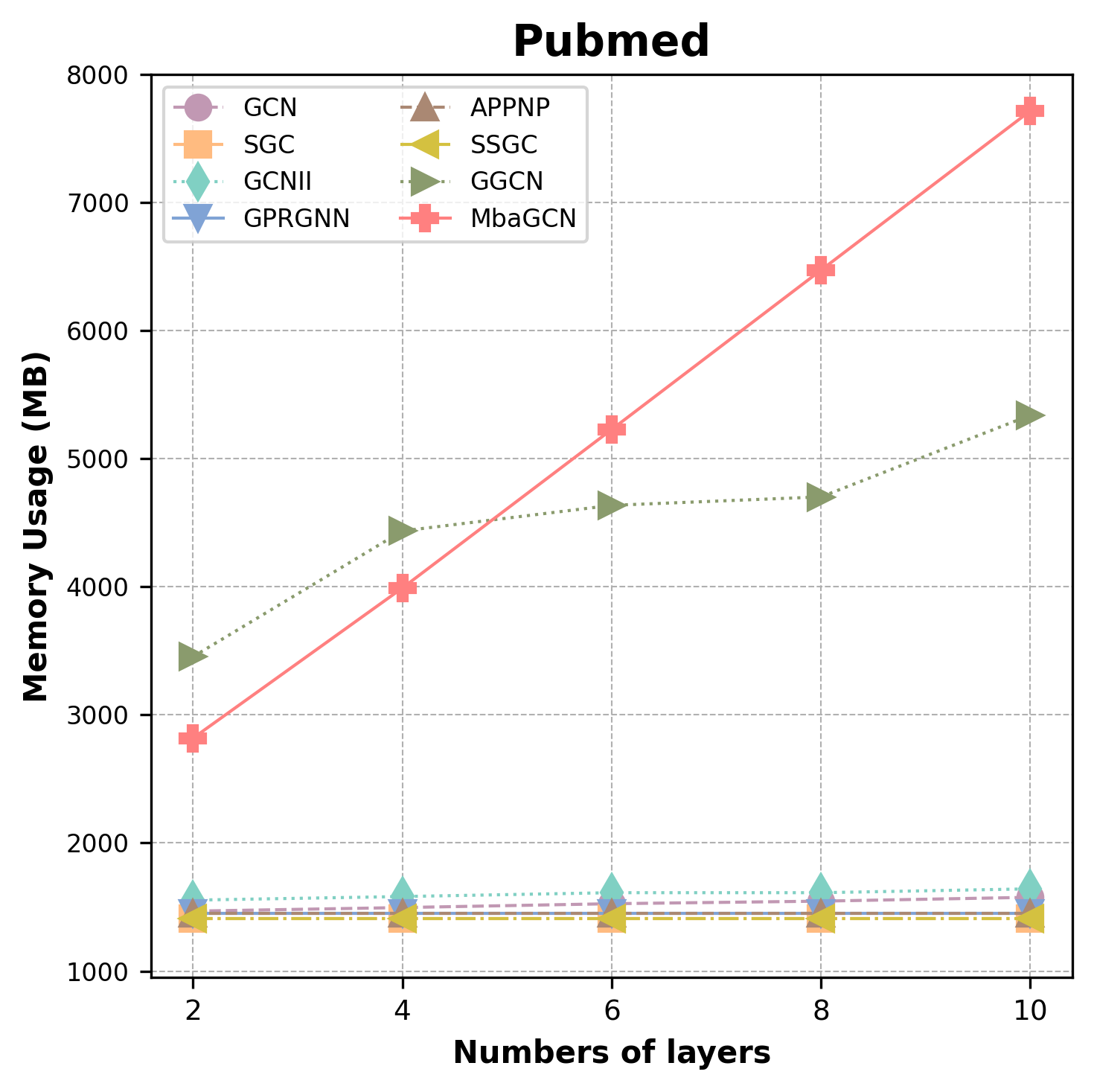}}}}%

\vskip -0.1in
\caption{Performance of baselines and the proposed MbaGCN with 2/4/6/8/10 layers.} \label{fig:time_memory_cost}
\vskip -0.1in
\end{figure}

To further investigate the efficiency and scalability of our model, we conduct experiments to analyze the trends in time and memory usage as the number of layers increases. The results are then compared with all baseline methods. The result of the experiment is shown in Fig.~\ref{fig:time_memory_cost}.

From the figure, we can observe that as the number of layers increases, most baseline methods exhibit no significant growth in time and memory consumption. This is because these methods primarily increase the number of layers by adding simple aggregation operations without introducing additional learnable parameters, resulting in a relatively flat growth in both time and space consumption.
In contrast, our proposed MbaGCN exhibits a gradual increase in time consumption and a linear increase in memory consumption. This is due to the fact that the number of parameters in MbaGCN is proportional to the number of layers, meaning that as the number of layers increases, the model's parameter count increases accordingly, leading to a linear rise in memory usage. 

Although our model shows differences from the baseline methods in terms of time and memory consumption, it presents a novel approach by solving the over-smoothing problem via the selection state space model and improving the model's expressive capacity. Future work will focus on optimizing the algorithm to reduce time and memory consumption, further enhancing the model's potential for large-scale graph data applications.

%% file: ijcai25.bbl
\begin{thebibliography}{}

\bibitem[\protect\citeauthoryear{Behrouz and Hashemi}{2024}]{behrouz2024graph}
Ali Behrouz and Farnoosh Hashemi.
\newblock Graph mamba: Towards learning on graphs with state space models.
\newblock In {\em Proceedings of the 30th ACM SIGKDD Conference on Knowledge Discovery and Data Mining}, pages 119--130, 2024.

\bibitem[\protect\citeauthoryear{Chen \bgroup \em et al.\egroup }{2020}]{chen2020simple}
Ming Chen, Zhewei Wei, Zengfeng Huang, Bolin Ding, and Yaliang Li.
\newblock Simple and deep graph convolutional networks.
\newblock In {\em International conference on machine learning}, pages 1725--1735. PMLR, 2020.

\bibitem[\protect\citeauthoryear{Chien \bgroup \em et al.\egroup }{2020}]{chien2020adaptive}
Eli Chien, Jianhao Peng, Pan Li, and Olgica Milenkovic.
\newblock Adaptive universal generalized pagerank graph neural network.
\newblock {\em arXiv preprint arXiv:2006.07988}, 2020.

\bibitem[\protect\citeauthoryear{Dao and Gu}{2024}]{dao2024transformers}
Tri Dao and Albert Gu.
\newblock Transformers are ssms: Generalized models and efficient algorithms through structured state space duality.
\newblock {\em arXiv preprint arXiv:2405.21060}, 2024.

\bibitem[\protect\citeauthoryear{Ding \bgroup \em et al.\egroup }{2024}]{ding2024recurrent}
Yuhui Ding, Antonio Orvieto, Bobby He, and Thomas Hofmann.
\newblock Recurrent distance filtering for graph representation learning.
\newblock In {\em Forty-first International Conference on Machine Learning}, 2024.

\bibitem[\protect\citeauthoryear{Gasteiger \bgroup \em et al.\egroup }{2018}]{gasteiger2018predict}
Johannes Gasteiger, Aleksandar Bojchevski, and Stephan G{\"u}nnemann.
\newblock Predict then propagate: Graph neural networks meet personalized pagerank.
\newblock {\em arXiv preprint arXiv:1810.05997}, 2018.

\bibitem[\protect\citeauthoryear{Grazzi \bgroup \em et al.\egroup }{2024}]{grazzi2024mamba}
Riccardo Grazzi, Julien Siems, Simon Schrodi, Thomas Brox, and Frank Hutter.
\newblock Is mamba capable of in-context learning?
\newblock {\em arXiv preprint arXiv:2402.03170}, 2024.

\bibitem[\protect\citeauthoryear{Gu and Dao}{2023}]{gu2023mamba}
Albert Gu and Tri Dao.
\newblock Mamba: Linear-time sequence modeling with selective state spaces.
\newblock {\em arXiv preprint arXiv:2312.00752}, 2023.

\bibitem[\protect\citeauthoryear{Gu \bgroup \em et al.\egroup }{2020}]{gu2020hippo}
Albert Gu, Tri Dao, Stefano Ermon, Atri Rudra, and Christopher R{\'e}.
\newblock Hippo: Recurrent memory with optimal polynomial projections.
\newblock {\em Advances in neural information processing systems}, 33:1474--1487, 2020.

\bibitem[\protect\citeauthoryear{Gu \bgroup \em et al.\egroup }{2021}]{gu2021efficiently}
Albert Gu, Karan Goel, and Christopher R{\'e}.
\newblock Efficiently modeling long sequences with structured state spaces.
\newblock {\em arXiv preprint arXiv:2111.00396}, 2021.

\bibitem[\protect\citeauthoryear{He \bgroup \em et al.\egroup }{2021}]{he2021bernnet}
Mingguo He, Zhewei Wei, Hongteng Xu, et~al.
\newblock Bernnet: Learning arbitrary graph spectral filters via bernstein approximation.
\newblock {\em Advances in Neural Information Processing Systems}, 34:14239--14251, 2021.

\bibitem[\protect\citeauthoryear{Hu \bgroup \em et al.\egroup }{2025}]{hu2025zigma}
Vincent~Tao Hu, Stefan~Andreas Baumann, Ming Gui, Olga Grebenkova, Pingchuan Ma, Johannes Fischer, and Bj{\"o}rn Ommer.
\newblock Zigma: A dit-style zigzag mamba diffusion model.
\newblock In {\em European Conference on Computer Vision}, pages 148--166. Springer, 2025.

\bibitem[\protect\citeauthoryear{Jin \bgroup \em et al.\egroup }{2021}]{jin2021node}
Wei Jin, Tyler Derr, Yiqi Wang, Yao Ma, Zitao Liu, and Jiliang Tang.
\newblock Node similarity preserving graph convolutional networks.
\newblock In {\em Proceedings of the 14th ACM international conference on web search and data mining}, pages 148--156, 2021.

\bibitem[\protect\citeauthoryear{Kipf and Welling}{2016}]{kipf2016semi}
Thomas~N Kipf and Max Welling.
\newblock Semi-supervised classification with graph convolutional networks.
\newblock {\em arXiv preprint arXiv:1609.02907}, 2016.

\bibitem[\protect\citeauthoryear{Li \bgroup \em et al.\egroup }{2018}]{li2018deeper}
Qimai Li, Zhichao Han, and Xiao-Ming Wu.
\newblock Deeper insights into graph convolutional networks for semi-supervised learning.
\newblock In {\em Proceedings of the AAAI conference on artificial intelligence}, volume~32, 2018.

\bibitem[\protect\citeauthoryear{Li \bgroup \em et al.\egroup }{2021}]{li2021dual}
Ruifan Li, Hao Chen, Fangxiang Feng, Zhanyu Ma, Xiaojie Wang, and Eduard Hovy.
\newblock Dual graph convolutional networks for aspect-based sentiment analysis.
\newblock In {\em Proceedings of the 59th Annual Meeting of the Association for Computational Linguistics and the 11th International Joint Conference on Natural Language Processing (Volume 1: Long Papers)}, pages 6319--6329, 2021.

\bibitem[\protect\citeauthoryear{Lieber \bgroup \em et al.\egroup }{2024}]{lieber2024jamba}
Opher Lieber, Barak Lenz, Hofit Bata, Gal Cohen, Jhonathan Osin, Itay Dalmedigos, Erez Safahi, Shaked Meirom, Yonatan Belinkov, Shai Shalev-Shwartz, et~al.
\newblock Jamba: A hybrid transformer-mamba language model.
\newblock {\em arXiv preprint arXiv:2403.19887}, 2024.

\bibitem[\protect\citeauthoryear{Miao \bgroup \em et al.\egroup }{2024}]{miao2024rethinking}
Rui Miao, Kaixiong Zhou, Yili Wang, Ninghao Liu, Ying Wang, and Xin Wang.
\newblock Rethinking independent cross-entropy loss for graph-structured data.
\newblock In {\em Proceedings of the 41st International Conference on Machine Learning}, pages 35570--35589, 2024.

\bibitem[\protect\citeauthoryear{Qin \bgroup \em et al.\egroup }{2024}]{qin2024learning}
Yifang Qin, Wei Ju, Hongjun Wu, Xiao Luo, and Ming Zhang.
\newblock Learning graph ode for continuous-time sequential recommendation.
\newblock {\em IEEE Transactions on Knowledge and Data Engineering}, 2024.

\bibitem[\protect\citeauthoryear{Rusch \bgroup \em et al.\egroup }{2023}]{rusch2023survey}
T~Konstantin Rusch, Michael~M Bronstein, and Siddhartha Mishra.
\newblock A survey on oversmoothing in graph neural networks.
\newblock {\em arXiv preprint arXiv:2303.10993}, 2023.

\bibitem[\protect\citeauthoryear{Shen \bgroup \em et al.\egroup }{2024a}]{shen2024graph}
Xu~Shen, Pietro Lio, Lintao Yang, Ru~Yuan, Yuyang Zhang, and Chengbin Peng.
\newblock Graph rewiring and preprocessing for graph neural networks based on effective resistance.
\newblock {\em IEEE Transactions on Knowledge and Data Engineering}, 2024.

\bibitem[\protect\citeauthoryear{Shen \bgroup \em et al.\egroup }{2024b}]{shen2024optimizing}
Xu~Shen, Yili Wang, Kaixiong Zhou, Shirui Pan, and Xin Wang.
\newblock Optimizing ood detection in molecular graphs: A novel approach with diffusion models.
\newblock In {\em Proceedings of the 30th ACM SIGKDD Conference on Knowledge Discovery and Data Mining}, pages 2640--2650, 2024.

\bibitem[\protect\citeauthoryear{Shen \bgroup \em et al.\egroup }{2025}]{shen2025raising}
Xu~Shen, Yixin Liu, Yili Wang, Rui Miao, Yiwei Dai, Shirui Pan, and Xin Wang.
\newblock Raising the bar in graph ood generalization: Invariant learning beyond explicit environment modeling.
\newblock {\em arXiv preprint arXiv:2502.10706}, 2025.

\bibitem[\protect\citeauthoryear{Tang \bgroup \em et al.\egroup }{2023}]{tang2023modeling}
Siyi Tang, Jared~A Dunnmon, Qu~Liangqiong, Khaled~K Saab, Tina Baykaner, Christopher Lee-Messer, and Daniel~L Rubin.
\newblock Modeling multivariate biosignals with graph neural networks and structured state space models.
\newblock In {\em Conference on Health, Inference, and Learning}, pages 50--71. PMLR, 2023.

\bibitem[\protect\citeauthoryear{Veli{\v{c}}kovi{\'c} \bgroup \em et al.\egroup }{2017}]{velivckovic2017graph}
Petar Veli{\v{c}}kovi{\'c}, Guillem Cucurull, Arantxa Casanova, Adriana Romero, Pietro Lio, and Yoshua Bengio.
\newblock Graph attention networks.
\newblock {\em arXiv preprint arXiv:1710.10903}, 2017.

\bibitem[\protect\citeauthoryear{Wan \bgroup \em et al.\egroup }{2021}]{wan2021contrastive}
Sheng Wan, Shirui Pan, Jian Yang, and Chen Gong.
\newblock Contrastive and generative graph convolutional networks for graph-based semi-supervised learning.
\newblock In {\em Proceedings of the AAAI conference on artificial intelligence}, volume~35, pages 10049--10057, 2021.

\bibitem[\protect\citeauthoryear{Wang \bgroup \em et al.\egroup }{2021}]{wang2021structure}
Bo~Wang, Tao Shen, Guodong Long, Tianyi Zhou, Ying Wang, and Yi~Chang.
\newblock Structure-augmented text representation learning for efficient knowledge graph completion.
\newblock In {\em Proceedings of the Web Conference 2021}, pages 1737--1748, 2021.

\bibitem[\protect\citeauthoryear{Wang \bgroup \em et al.\egroup }{2022}]{wang2022adagcl}
Yili Wang, Kaixiong Zhou, Rui Miao, Ninghao Liu, and Xin Wang.
\newblock Adagcl: Adaptive subgraph contrastive learning to generalize large-scale graph training.
\newblock In {\em Proceedings of the 31st ACM international conference on information \& knowledge management}, pages 2046--2055, 2022.

\bibitem[\protect\citeauthoryear{Wang \bgroup \em et al.\egroup }{2024a}]{wang2024graph}
Chloe Wang, Oleksii Tsepa, Jun Ma, and Bo~Wang.
\newblock Graph-mamba: Towards long-range graph sequence modeling with selective state spaces.
\newblock {\em arXiv preprint arXiv:2402.00789}, 2024.

\bibitem[\protect\citeauthoryear{Wang \bgroup \em et al.\egroup }{2024b}]{wang2024unifying}
Yili Wang, Yixin Liu, Xu~Shen, Chenyu Li, Kaize Ding, Rui Miao, Ying Wang, Shirui Pan, and Xin Wang.
\newblock Unifying unsupervised graph-level anomaly detection and out-of-distribution detection: A benchmark.
\newblock {\em arXiv preprint arXiv:2406.15523}, 2024.

\bibitem[\protect\citeauthoryear{Wang \bgroup \em et al.\egroup }{2024c}]{GraphSAM}
Yili Wang, Kaixiong Zhou, Ninghao Liu, Ying Wang, and Xin Wang.
\newblock Efficient sharpness-aware minimization for molecular graph transformer models.
\newblock In {\em The Twelfth International Conference on Learning Representations, {ICLR} 2024, Vienna, Austria, May 7-11, 2024}. OpenReview.net, 2024.

\bibitem[\protect\citeauthoryear{Wang \bgroup \em et al.\egroup }{2025}]{wang2025mamba}
Zihan Wang, Fanheng Kong, Shi Feng, Ming Wang, Xiaocui Yang, Han Zhao, Daling Wang, and Yifei Zhang.
\newblock Is mamba effective for time series forecasting?
\newblock {\em Neurocomputing}, 619:129178, 2025.

\bibitem[\protect\citeauthoryear{Wu \bgroup \em et al.\egroup }{2019}]{wu2019simplifying}
Felix Wu, Amauri Souza, Tianyi Zhang, Christopher Fifty, Tao Yu, and Kilian Weinberger.
\newblock Simplifying graph convolutional networks.
\newblock In {\em International conference on machine learning}, pages 6861--6871. PMLR, 2019.

\bibitem[\protect\citeauthoryear{Wu \bgroup \em et al.\egroup }{2024}]{wu2024demystifying}
Xinyi Wu, Amir Ajorlou, Zihui Wu, and Ali Jadbabaie.
\newblock Demystifying oversmoothing in attention-based graph neural networks.
\newblock {\em Advances in Neural Information Processing Systems}, 36, 2024.

\bibitem[\protect\citeauthoryear{Xu \bgroup \em et al.\egroup }{2021}]{xu2021self}
Minghao Xu, Hang Wang, Bingbing Ni, Hongyu Guo, and Jian Tang.
\newblock Self-supervised graph-level representation learning with local and global structure.
\newblock In {\em International Conference on Machine Learning}, pages 11548--11558. PMLR, 2021.

\bibitem[\protect\citeauthoryear{Yan \bgroup \em et al.\egroup }{2022}]{yan2022two}
Yujun Yan, Milad Hashemi, Kevin Swersky, Yaoqing Yang, and Danai Koutra.
\newblock Two sides of the same coin: Heterophily and oversmoothing in graph convolutional neural networks.
\newblock In {\em 2022 IEEE International Conference on Data Mining (ICDM)}, pages 1287--1292. IEEE, 2022.

\bibitem[\protect\citeauthoryear{Yang \bgroup \em et al.\egroup }{2024}]{yang2024plainmamba}
Chenhongyi Yang, Zehui Chen, Miguel Espinosa, Linus Ericsson, Zhenyu Wang, Jiaming Liu, and Elliot~J Crowley.
\newblock Plainmamba: Improving non-hierarchical mamba in visual recognition.
\newblock {\em arXiv preprint arXiv:2403.17695}, 2024.

\bibitem[\protect\citeauthoryear{Yu \bgroup \em et al.\egroup }{2022}]{yu2022multiplex}
Pengyang Yu, Chaofan Fu, Yanwei Yu, Chao Huang, Zhongying Zhao, and Junyu Dong.
\newblock Multiplex heterogeneous graph convolutional network.
\newblock In {\em Proceedings of the 28th ACM SIGKDD Conference on Knowledge Discovery and Data Mining}, pages 2377--2387, 2022.

\bibitem[\protect\citeauthoryear{Yuan \bgroup \em et al.\egroup }{2024}]{yuan2024instance}
Xiaosong Yuan, Chen Shen, Shaotian Yan, Xiaofeng Zhang, Liang Xie, Wenxiao Wang, Renchu Guan, Ying Wang, and Jieping Ye.
\newblock Instance-adaptive zero-shot chain-of-thought prompting.
\newblock {\em arXiv preprint arXiv:2409.20441}, 2024.

\bibitem[\protect\citeauthoryear{Zhai \bgroup \em et al.\egroup }{2024}]{zhai2024bregman}
Jiayu Zhai, Lequan Lin, Dai Shi, and Junbin Gao.
\newblock Bregman graph neural network.
\newblock In {\em ICASSP 2024-2024 IEEE International Conference on Acoustics, Speech and Signal Processing (ICASSP)}, pages 6250--6254. IEEE, 2024.

\bibitem[\protect\citeauthoryear{Zhang and Li}{2021}]{zhang2021nested}
Muhan Zhang and Pan Li.
\newblock Nested graph neural networks.
\newblock {\em Advances in Neural Information Processing Systems}, 34:15734--15747, 2021.

\bibitem[\protect\citeauthoryear{Zhang \bgroup \em et al.\egroup }{2021}]{zhang2021node}
Wentao Zhang, Mingyu Yang, Zeang Sheng, Yang Li, Wen Ouyang, Yangyu Tao, Zhi Yang, and Bin Cui.
\newblock Node dependent local smoothing for scalable graph learning.
\newblock {\em Advances in Neural Information Processing Systems}, 34:20321--20332, 2021.

\bibitem[\protect\citeauthoryear{Zhu and Koniusz}{2021}]{zhu2021simple}
Hao Zhu and Piotr Koniusz.
\newblock Simple spectral graph convolution.
\newblock In {\em International conference on learning representations}, 2021.

\bibitem[\protect\citeauthoryear{Zhu \bgroup \em et al.\egroup }{2024}]{zhu2024vision}
Lianghui Zhu, Bencheng Liao, Qian Zhang, Xinlong Wang, Wenyu Liu, and Xinggang Wang.
\newblock Vision mamba: Efficient visual representation learning with bidirectional state space model.
\newblock {\em arXiv preprint arXiv:2401.09417}, 2024.

\end{thebibliography}
